\pdfoutput=1
\documentclass[10pt,twocolumn,letterpaper]{article}
\usepackage{cvpr}              % To produce the CAMERA-READY version
\usepackage[accsupp]{axessibility}
\usepackage{graphicx}
\usepackage{amsmath}
\usepackage{amssymb}
\usepackage{booktabs}
\usepackage{float}
\usepackage{cuted}
\usepackage{capt-of}
\usepackage{multirow}
\usepackage{bbding}
\usepackage[pagebackref,breaklinks,colorlinks]{hyperref}

\usepackage[capitalize]{cleveref}
\crefname{section}{Sec.}{Secs.}
\Crefname{section}{Section}{Sections}
\Crefname{table}{Table}{Tables}
\crefname{table}{Tab.}{Tabs.}

\def\cvprPaperID{2562} % *** Enter the CVPR Paper ID here
\def\confName{CVPR}
\def\confYear{2022}

\begin{document}

\title{Optimizing Video Prediction via Video Frame Interpolation}

\author{
Yue Wu \qquad Qiang Wen \qquad Qifeng Chen\\
The Hong Kong University of Science and Technology\\ }
\maketitle

\begin{strip}
\centering
\includegraphics[width=1.0\linewidth]{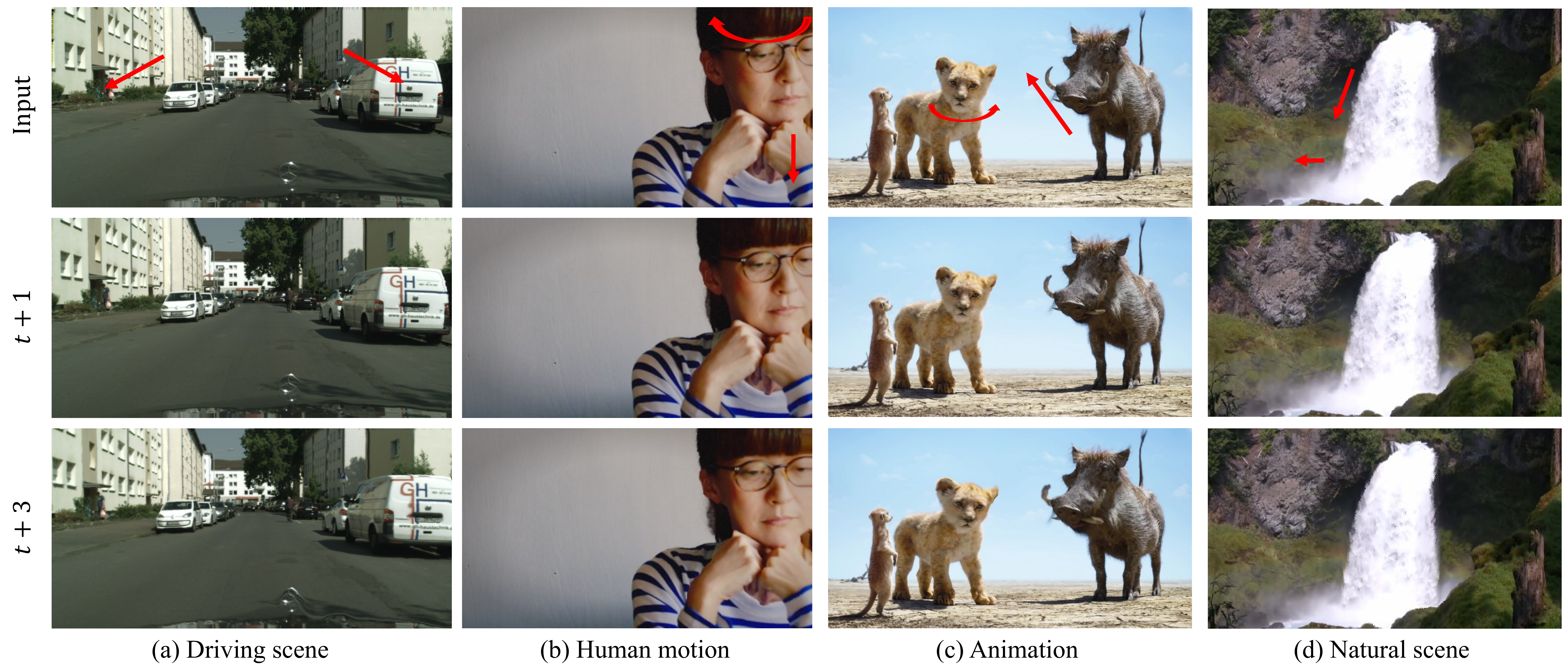}
\captionof{figure}{Our method can produce plausible video prediction results in diverse scenarios without external training, such as driving scenes, human motion, animation, and natural scenes. We use red arrows to indicate the motions from input frames. Video results are presented in the supplementary material. \label{fig:teaser}}
\end{strip}

%%%%%%%%% ABSTRACT
\begin{abstract}
Video prediction is an extrapolation task that predicts future frames given past frames, and video frame interpolation is an interpolation task that estimates intermediate frames between two frames. We have witnessed the tremendous advancement of video frame interpolation, but the general video prediction in the wild is still an open question. Inspired by the photo-realistic results of video frame interpolation, we present a new optimization framework for video prediction via video frame interpolation, in which we solve an extrapolation problem based on an interpolation model. Our video prediction framework is based on optimization with a pretrained differentiable video frame interpolation module without the need for a training dataset, and thus there is no domain gap issue between training and test data. Also, our approach does not need any additional information such as semantic or instance maps, which makes our framework applicable to any video. Extensive experiments on the Cityscapes, KITTI, DAVIS, Middlebury, and Vimeo90K datasets show that our video prediction results are robust in general scenarios, and our approach outperforms other video prediction methods that require a large amount of training data or extra semantic information. 
\end{abstract}

%%%%%%%%% BODY TEXT
\section{Introduction}
\label{sec:intro}

Video prediction is an extrapolation task to predict future video frames given some past frames. Video prediction has broad applications including robotics planning, autonomous driving and video manipulations~\cite{bair_pushing,Srivastava2015,NextSegmPredICCV17,highfideity}. For instance, predicted videos can help autonomous robots to better plan future actions with future visual information. Video prediction is also a fundamental task for unconditional video synthesis that can be decomposed into image synthesis and future video prediction.

\begin{figure*}[t!]
\centering
\vspace{-2mm}
\includegraphics[width=0.995\linewidth]{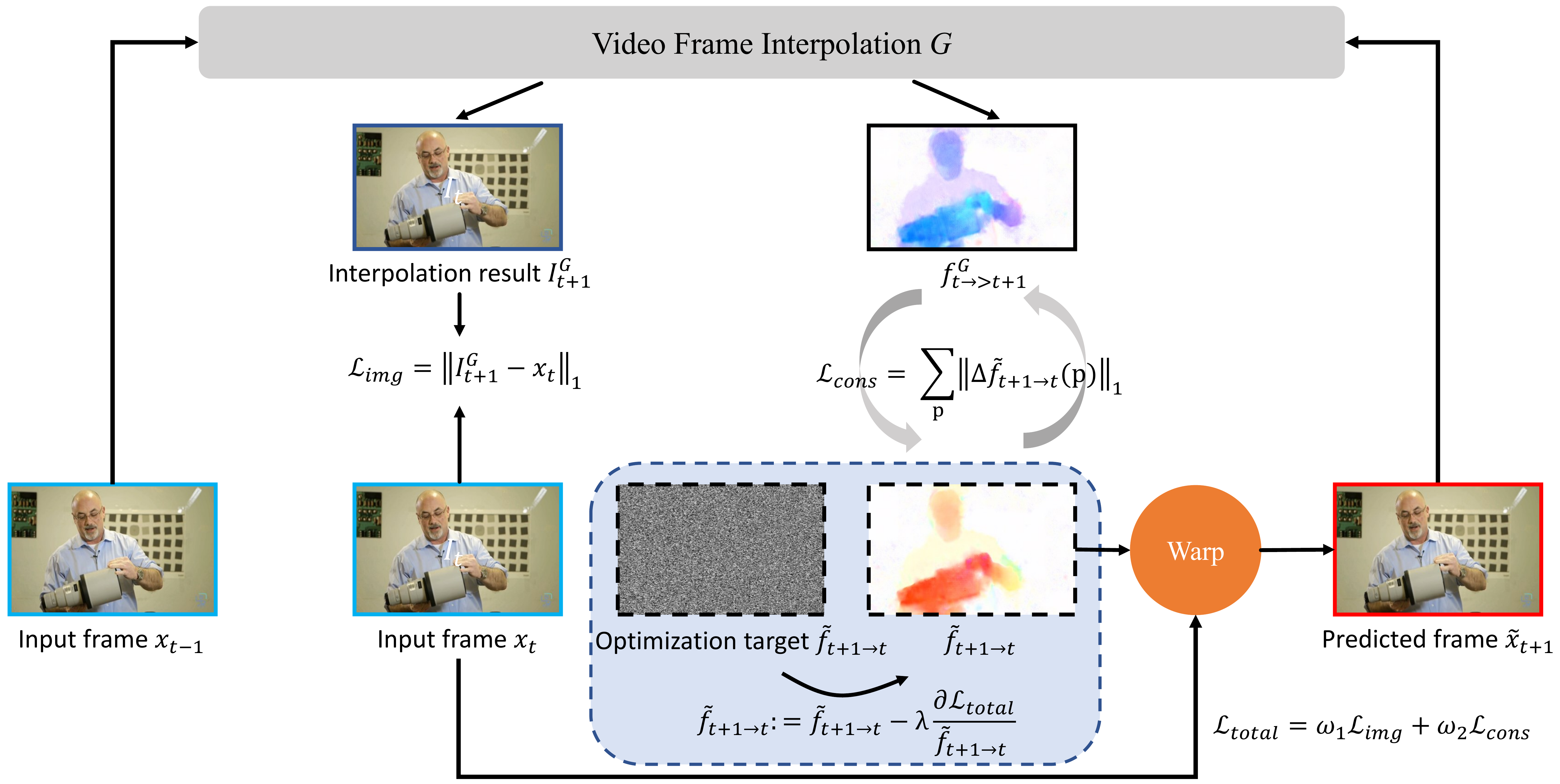}
\vspace{-2mm}
\caption{Overview of our method. We optimize optical flow $\tilde{f}_{t+1\rightarrow t}$ by a video frame interpolation $G$ \cite{RIFE}. Our optimization objective is image-level distance $\mathcal L_{img}$ and a consistency constraint between our predicted flow $\tilde{f}_{t+1\rightarrow t}$ and the flow $f^{G}_{t\rightarrow t+1}$ generated by $G$.}
\label{fig:overview}
\vspace{-2mm}
\end{figure*}

Video prediction is a challenging extrapolation problem. Several studies~\cite{prednet, villegas17mcnet, liu2017voxelflow, dyan} only take RGB frames as input for video prediction and find that the video prediction problem is difficult to solve, because of the inherent high complexity of  video prediction and the uncertainty of future states. Thus, recently, a lot of constraints and assumptions about the modeling scene have been employed to simplify the problem. However, these assumptions reduce the generalization ability of these video prediction models. FVS \cite{FVS} and Lee \etal~
\cite{revisit21} require semantic maps to decompose the scene. Bei~\etal~\cite{sadm} first predict semantic maps and then synthesize future frames. Qi~\etal~\cite{Qi2019} require depth maps to reconstruct 3D point clouds. However, such additional information is often hard to obtain or estimate correctly in general scenarios. These strong assumptions limit these methods to be only applicable to data when these assumptions hold. For example, failing to detect some objects (including unseen objects) will lead to performance degradation. When this extra information is unavailable or of poor quality, these methods suffer from performance degradation and cannot be even applied in diverse videos in the real world.
%Given the diversity of video sequences in the real world, there exists a domain gap between training and testing.   

Moreover, these external methods usually need to train one model for each specific scenario, making them difficult generalize to other scenarios. For example, it is hard to apply a video prediction model trained on a driving scene to a human moving dataset, as the motion difference is huge. 

To address these issues, we propose an optimization-based video prediction method, without the requirement of external training (no external dataset is needed for training), and can produce state-of-the-art results. Our insight is that we can cast the video prediction problem as a video frame interpolation (VFI) based optimization problem. Inspired by the recent success of VFI, we connect these two problems to solve the video prediction problem in a new way. Our method does not require any assumption such as semantic segmentation and can be applied to any video.
%The predicted future state should be able to interpolate back to the intermediate state. 
We evaluate our method on multiple datasets, and our method can outperform the state of the arts.
Our contributions can be summarized as follows:
\begin{itemize}
     \item We present the first optimization framework for video prediction. We cast the extrapolation problem of video prediction as an optimization problem with VFI. 
    \item Our framework is highly flexible as it does not require any semantic or instance maps, prior knowledge about the scene, or external training. Our method is applicable to video prediction in any scene at any resolution.
    \item Our method obtains outstanding performance on various datasets and outperforms state-of-the-art video prediction approaches that require additional information. Our method surpasses external learning methods taking only RGB frames as input by a large margin.
\end{itemize}

\section{Related work}
\subsection{Video Prediction}
% The objective of video prediction is to synthesize future frames based on observed past video sequences.  
% There are two streams of video prediction frameworks. The first one is stochastic methods~\cite{denton2018stochastic} to explore the future uncertainty. The second one is deterministic methods, aiming at improving model capacity to predict deterministic changes better. Our paper addresses the second aspect. 
%Our paper addresses the deterministic video prediction problem, which aims at improving model capacity to predict deterministic changes better.
Early works, taking RGB frames as input, employ several mechanisms to improve video prediction.
MCNet~\cite{villegas17mcnet} decomposes a scene into content and motion components. 
DVF~\cite{liu2017voxelflow} proposes deep voxel flow to synthesize future frames. 
However, video prediction in the wild is still quite hard due to its inherent high complexity and uncertainty.
Thus, explicit modeling, constraints, and assumptions about the scene are introduced. Qi~\etal~\cite{Qi2019} utilize depth maps to reconstruct 3D point clouds. Gao \etal~\cite{DPG} use semantic maps to enforce layout consistency. FVS~\cite{FVS} needs semantic segmentation and instance segmentation to decompose a scene into background and foreground identities. Bei~\etal~\cite{sadm} and Lee~\etal~\cite{revisit21} also require semantic maps. 

Although the performance of video prediction has been gradually improved, the generalization ability of these approaches is arguably reduced. It is hard to apply these methods to data without these extra annotations. Moreover, these methods may suffer from performance degradation when the test data is from a different domain than the train data. For example, it is hard to apply a model trained on robot scenarios to driving scenes because the motion in these two domains is quite different. Thus, we propose an optimization-based video prediction method, which can produce state-of-the-art results without external training (thus no domain gap).

\subsection{Video Frame Interpolation}
Contrary to the challenges of video prediction problems, VFI has gained great success recently, which motivates our model design. VFI aims to interpolate intermediate frames between successive input frames. There are three categories of algorithms: kernel-based \cite{niklaus2017adaconv, niklaus2017sepconv, bao2018memc, bao2019dain}, phase-based \cite{meyer2015phase, meyer2018phase}, and motion-based \cite{liu2017dvf, liu2019cyclicgen, bao2018memc, bao2019dain, Super_SloMo, niklaus2018ctx, gui2020feflow, park2020bmbc, niklaus2020softsplatting,park2021ABME}. 
These motion-based methods use bi-directional optical flows to warp two consecutive frames forward and backward to obtain an intermediate frame. Our method adopts a motion-based framework. Super SloMo~\cite{Super_SloMo} linearly combines bi-directional flows as an initial approximation of the intermediate flows for further refinement. Park~\etal~\cite{park2021ABME} conduct asymmetric bilateral motion estimation. Sim~\etal~\cite{XVFI} first handle the VFI for 4K videos with large motion.
%Xu \etal~\cite{qvi_nips19} exploits flow reversal filter to produce the intermediate flows.
RIFE~\cite{RIFE} is a real-time interpolation algorithm that estimates the intermediate flows in a coarse-to-fine fashion.
Different from the domain gap problem in video prediction, there are much fewer domain constraints in VFI. These methods do not require additional information, such as semantic maps and depth maps, and can produce outstanding interpolation performance even with complex motion. Inspired by the success of VFI, we cast video prediction as an optimization problem based on VFI.

\subsection{Optimization-based Methods}
Since learning-based methods suffer from the domain gap between training data and test data, optimization-based methods on test data are still competitive nowadays. 
Gatys \etal \cite{gatys2016image} propose the first optimization-based method for neural style transfer. 
% Given a content image and a style image, Gatys \etal compute the deep representations of content and style images and optimize for an image with style transfer via a deep feature loss and a style loss. 
%Based on the external prior from the pretrained VGG-19 network \cite{simonyan2014very}, this method not only presents a satisfying style transfer performance but also remains the natural clues in the original content image. 
Shaham \etal \cite{shaham2019singan} optimize a generative model on a single image and can generate high quality and diverse samples from the image. 
Lei~\etal~\cite{lei2020dvp, DVP_lei} optimize a network to improve the temporal consistency.
Mildenhall \etal \cite{mildenhall2020nerf} optimize a fully-connected network on a sparse set of input views from a scene for novel view synthesis. 
%Their method represents the 3D scene as a neural radiance field and can render novel views. 
These optimization-based methods inspire us to propose a framework that not only tackles the domain gap problem but also provides an on-the-fly control for users during optimization.

\section{Method}
\subsection{Problem Formulation}
Let $x_t$ be the video frame at time step $t$. The input to our framework includes two recent RGB frames $x_{t-1}$ and $x_{t}$. Our goal is to predict the future frames $\{\tilde{x}_{t+1},\tilde{x}_{t+2},\hdots\}$. We adopt a pretrained video interpolation network~\cite{RIFE} denoted as $G$. 
%We cast the video prediction problem as an optimization problem. 
We will focus on predicting the next frame $\tilde{x}_{t+1}$ first as we can predict future frames one by one sequentially. During the optimization process, the parameters of $G$ are unchanged. Our primary objective is
\begin{equation}\label{eq1}
     \tilde{x}_{t+1}^{*} = \underset{\tilde{x}_{t+1}}{\mathrm{argmin}}\, E(G(x_{t-1}, \tilde{x}_{t+1}), x_{t}),
\end{equation}
\noindent where $E$ is an objective function that measures image similarity. Here we utilize a VFI network $G$ to constrain the relationship among $x_{t-1}$, $\tilde{x}_{t+1}$, and $x_{t}$.

To ease the optimization process, we choose to optimize optical flow $\tilde{f}_{t+1 \rightarrow t}$ between the predicted frame $\tilde{x}_{t+1}$ and the last observed frame $x_t$ instead of directly optimize $\tilde{x}_{t+1}$.  $\tilde{x}_{t+1}$ is computed using backward warping~\cite{jaderberg2015spatial}:
\begin{equation}
    \tilde{x}_{t+1} = warp(x_t, \tilde{f}_{t+1 \rightarrow t}).
\end{equation}\label{warp}
Then Eq.~\ref{eq1} can be rewritten as
\begin{equation}\label{eq2}
\hspace{-2mm}
     {\tilde{f}_{t+1 \rightarrow t}} ^{*} = \underset{\tilde{f}_{t+1 \rightarrow t}}{\mathrm{argmin}}\, E(G(x_{t-1},  warp(x_t, \tilde{f}_{t+1 \rightarrow t})), x_{t}).
\end{equation}

\paragraph{Flow initialization.} To ease the optimization of Eq.~\ref{eq2}, it is a good practice to start with a flow $\tilde{f}_{t+1 \rightarrow t}$ that produces an approximate motion. Therefore, we initialize it utilizing the negative flow of $f_{t \rightarrow t-1}$:
\begin{equation}
    \tilde{f}_{t+1 \rightarrow t} = \delta(- f_{t \rightarrow t-1}).
\end{equation}
We first compute $- f_{t \rightarrow t-1}$ as a rough approximation of $f_{t->t+1}$. 
Then we initialize $\tilde{f}_{t+1 \rightarrow t}$ as the inversion of $- f_{t \rightarrow t-1}$. 
$\delta$ represents the operation similar to the flow reversal layer~\cite{qvi_nips19} to convert a forward flow to a backward flow (details in the supplement).

However, directly optimizing Eq.~\ref{eq2} is still difficult, because the constraint towards $\tilde{f}_{t+1 \rightarrow t}$ is indirect, and the optimization process is difficult to converge. 
% It is difficult to drag $\tilde{f}_{t+1 \rightarrow t}$ towards the correct optimization direction. 

\subsection{Video Frame Interpolation Network} 
Thus, we propose to utilize the intermediate results of network $G$. Given $x_{t-1}$ and $\tilde{x}_{t+1}$ as input, $G$ generates optical flows of two directions $f^G_{t \rightarrow t-1}$, $f^G_{t \rightarrow t+1}$, and a mask $m^G$. The superscript $G$ denotes that it is the output of network $G$. The video interpolation network warps $x_{t-1}$ and $\tilde{x}_{t+1}$ towards time step $t$:
\begin{eqnarray}
   & I^G_{t-1} & =  warp(x_{t-1}, f^G_{t \rightarrow t-1}), \\
   & I^G_{t+1} & =  warp(\tilde{x}_{t+1}, f^G_{t \rightarrow t+1}), \\
   & I^G_t & =  I^G_{t-1} \times m^G + I^G_{t+1} \times ( 1 - m^G),
\end{eqnarray}
where $I^G_{t-1}$ is the intermediate interpolation frame by warping $x_{t-1}$ using $f^G_{t \rightarrow t-1}$. $I^G_{t+1}$ is the intermediate interpolation frame by warping $\tilde{x}_{t+1}$ using $f^G_{t \rightarrow t+1}$. The final interpolation result is a weighted sum of $I^G_{t-1}$ and $I^G_{t-1}$. 

We utilize $I^G_{t+1}$ instead of $I^G_t$ since $I^G_{t+1}$ has a closer relation with $\tilde{x}_{t+1}$ and can dismiss the effect of $m^G$. We employ a $L_1$ distance between $I^G_{t+1}$ and $x_t$:
\begin{equation}\label{eq:img}
    \mathcal{L}_{img} = \left \|I^G_{t+1}- x_t  \right \|_1.
\end{equation}
We also argue that there is a forward-backward consistency relationship between $f^G_{t \rightarrow t+1}$ and $\tilde{f}_{t+1 \rightarrow t}$. This constraint means that after the forward and backward propagation, the pixels should go back to the original locations:
\begin{equation}
    \mathcal{L}_{cons} = \sum_{\mathbf{p}} \left \| \Delta \tilde{f}_{{t+1}\rightarrow t }(\mathbf{p}) \right \|_1, 
\end{equation}
where $\Delta \tilde{f}_{{t+1}\rightarrow t }(\mathbf{p})$ is the discrepancy obtained from forward and backward flow check at pixel location $\mathbf{p}$:
\begin{eqnarray}
\Delta \Tilde{f}_{t+1 \rightarrow t}(\mathbf{p}) & = & \mathbf{p}- \big(\mathbf{p}' + f^G_{t \rightarrow t+1}(\mathbf{p}')\big), \\
\mathbf{p}' & = & \mathbf{p} + \Tilde{f}_{t+1 \rightarrow t}(\mathbf{p}).
\end{eqnarray}

Our overall objective function is $\mathcal{L}_{total} = \omega_1\mathcal{L}_{img} + \omega_2\mathcal{L}_{cons}$, where $\omega_1$ and $\omega_2$ are the loss weights.

\subsection{Flow Inpainting}\label{inpainting}
There always exist occlusion areas in optical flow: some pixels do not have corresponding pixels in successive frames. The estimated optical flow in occlusion areas is unreliable. Thus, we set a threshold $\alpha$ to mask out these areas, and using flow inpainting to fill in the holes:
\begin{equation}
\phi(\mathbf{p}) = \left\{\begin{matrix}
  &  1 & \quad if \quad \left\| \Delta \tilde{f}_{t+1 \rightarrow t}(\mathbf{p}) \right\|_1 > \alpha, \\
  &  0 & \quad otherwise.
\end{matrix}\right.
\end{equation}
%\noindent where $a$ is set as 1.5 empirically. 
If $\phi(\mathbf{p})$ is 1, its optical flow is treated unreliable, and is inpainted by a linear combination of neighboring valid flow values, whose weights are inversely proportional to the distance between the invalid pixel and the valid pixels.

We try to use adaptive weights like Softmax Splatting~\cite{niklaus2020softsplatting} to resolve the occlusions and find it unsuitable for our framework. Detailed analysis is described in the supplement.

\begin{table}[t!]
\vspace{-6mm}
\centering
\resizebox{0.998\linewidth}{!}{
\begin{tabular}{lcccc}
\toprule
\multicolumn{5}{c}{\textit{External learning methods}} \\
\midrule
   & External training & Semantic & Instance & Depth \\
\midrule
PredNet~\cite{prednet} & \checkmark & $\times$ & $\times$ & $\times$     \\
MCNET~\cite{villegas17mcnet} & \checkmark & $\times$ & $\times$ & $\times$   \\
DVF~\cite{liu2017voxelflow} & \checkmark & $\times$ & $\times$ & $\times$ \\
Vid2vid~\cite{Wang2018} & \checkmark & \checkmark & $\times$ & $\times$     \\
Qi \etal~\cite{Qi2019} & \checkmark & \checkmark & $\times$ & \checkmark \\
Seg2vid~\cite{seg2vid} & \checkmark & \checkmark & $\times$ & $\times$ \\ 
FVS~\cite{FVS}      & \checkmark & \checkmark & \checkmark & $\times$\\
HVP~\cite{revisit21} & \checkmark & \checkmark & $\times$ & $\times$\\
SADM~\cite{sadm} & \checkmark & \checkmark & $\times$ & $\times$\\
\hline
\multicolumn{5}{c}{\textit{Optimization methods}} \\
\hline
Ours & $\times$ & $\times$ & $\times$ & $\times$ \\
\bottomrule
\end{tabular}
}
\vspace{-1mm}
\caption{Comparison with other video prediction methods in terms of methods' requirements.}
\label{table:elements}
\vspace{-3mm}
\end{table}

\subsection{Implementation}

\paragraph{Multi-frame prediction.}
For multi-frame prediction, we choose to optimize the next frame recurrently. We try to set the optimization target as multiple optical flows and optimize multiple future frames together. The result turns out recurrently optimizing the next frame is more stable.

The input frame length for all datasets is set to 2. The hyperparameters $\omega_1$, $\omega_2$, $\alpha$ are set to 1.0, 3.0, 1.5 empirically. We adopt RAFT~\cite{RAFT} to estimate optical flow used for optimization target initialization.
%Our method is implemented with PyTorch. 
The Adam optimizer~\cite{kingma2014adam} is used with a learning rate of 0.1 for 3000 iterations for each future frame. We adopt the VFI method RIFE~\cite{RIFE}, which is pretrained only on Vimeo90K~\cite{vimeo}, for all the experiments. During the process of optimization, the network weight of RIFE~\cite{RIFE} is fixed.

\begin{table*}[t]
\vspace{-1mm}
\centering
\resizebox{1\linewidth}{!}{
\begin{tabular}{cccccccccccccc}
\toprule
&  & \multicolumn{6}{c}{Cityscapes} & \multicolumn{6}{c}{KITTI} \\
\midrule
& & \multicolumn{3}{c}{MS-SSIM ($\times1\mathrm{e}{-2}$)$\uparrow$} & \multicolumn{3}{c}{LPIPS ($\times1\mathrm{e}{-2}$)$\downarrow$} & \multicolumn{3}{c}{MS-SSIM ($\times1\mathrm{e}{-2}$)$\uparrow$} & 
\multicolumn{3}{c}{LPIPS ($\times1\mathrm{e}{-2}$)$\downarrow$}\\
& Input  & t+1 & t+3  & t+5 & t+1  & t+3 & t+5  & t+1 & t+3  & t+5 & t+1  & t+3 & t+5 \\ 
\midrule
 \multicolumn{14}{c}{\textit{External learning methods}} \\
\midrule
PredNet~\cite{prednet} & RGB & 84.03 & 79.25 & 75.21 & 25.99 & 29.99  & 36.03 & 56.26 &  51.47 & 47.56 & 55.35 & 58.66 & 62.95                \\
MCNET~\cite{villegas17mcnet} & RGB & 89.69 & 78.07 & 70.58 & 18.88 & 31.34 & 37.34 & 75.35  & 63.52 & 55.48 & 24.05  & 31.71 & 37.39 \\
DVF~\cite{liu2017voxelflow} & RGB & 83.85 & 76.23 & 71.11 & 17.37 & 24.05 & 28.79 & 53.93 & 46.99 & 42.62  & 32.47  & 37.43  & 41.59        \\
%DYAN~\cite{dyan} & RGB & 00.00 & 00.00 & 00.00 & 00.00 & 00.00 & 00.00 & 00.00 & 00.00 & 00.00 & 00.00  \\
Vid2vid~\cite{Wang2018} & RGB+S.       & 88.16 & 80.55 & 75.13 & 10.58 & 15.92 & 20.14  & N/A & N/A & N/A & N/A & N/A & N/A \\
Seg2vid~\cite{seg2vid} & RGB+S. & 88.32 & N/A & 61.63 & 9.69 & N/A & 25.99 & N/A & N/A & N/A& N/A& N/A& N/A\\ 
FVS~\cite{FVS}      & RGB+S.+I. & 89.10 & 81.13 & 75.68 & 8.50 & 12.98 & \textbf{16.50} & 79.28 & 67.65  &  60.77 & 18.48 & 24.61  &  30.49\\
\hline
 \multicolumn{14}{c}{\textit{Optimization methods}} \\
\hline
Ours & No external training & \textbf{94.54} & \textbf{86.89} & \textbf{80.40} & \textbf{6.46} & \textbf{12.50} & 17.83 & \textbf{82.71} & \textbf{69.50} & \textbf{61.09} & \textbf{12.34} & \textbf{20.29} & \textbf{26.35} \\
\bottomrule
\end{tabular}
}
\vspace{-1mm}
\caption{Comparison with state-of-the-art methods on the Cityscapes and KITTI datasets. S. and I. denote that the method requires semantic maps or instance maps as input. Our method can outperform previous video prediction methods by a large margin.}
\label{table:same}
\end{table*}

\begin{figure*}[h]
\vspace{-1mm}
\centering
\includegraphics[width=1\linewidth]{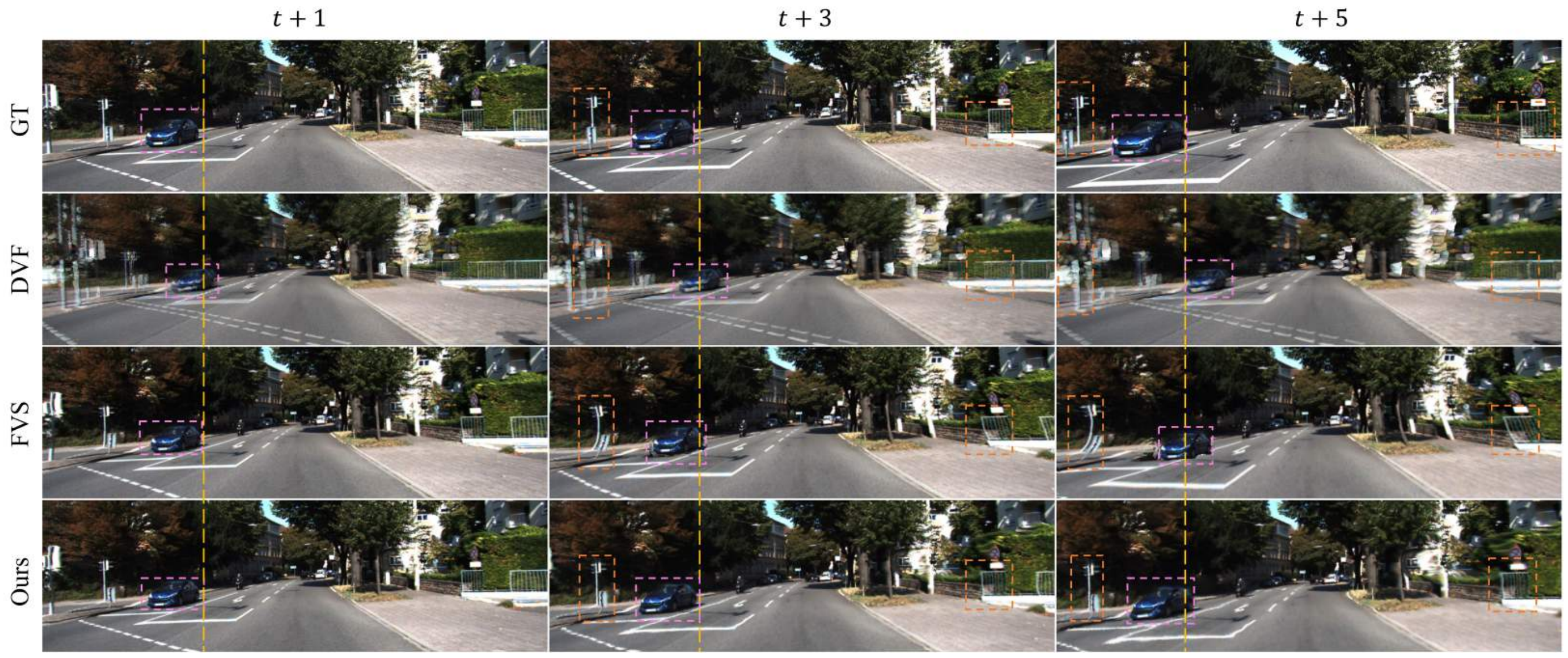}
\vspace{-4mm}
\caption{Multi-frame prediction comparison on KITTI. As the pink boxes show, DVF~\cite{liu2017voxelflow} fails to separate the motion of the blue car from the background and produces a ``zooming-in" effect, which is the major motion exhibited in the dataset, caused by the forward movement of the running car. FVS~\cite{FVS} wrongly predicts the motion of the blue car at $t+3$ and $t+5$, resulting in incorrect frame prediction results. Our model can correctly capture the motion of the blue car without external training. 
%Moreover, as depicted using orange boxes, our method can suppress distortion artifact by utilizing VFI. For example, to make the pole of the interpolation frame vertical, the pole of the predicted frame should also be vertical.
}
\label{fig:same}
\vspace{-3mm}
\end{figure*}

\section{Experiments}
We compare our method with state-of-the-art methods as shown in Table~\ref{table:elements} in terms of methods' requirements. Due to the complexity of future video prediction, many recent methods add some extra assumptions for video prediction, such as semantic maps~\cite{Wang2018,Qi2019,seg2vid,FVS,revisit21,sadm}, instance maps~\cite{FVS}, and depth maps~\cite{Qi2019}. These assumptions may improve prediction performance but vastly decrease their generalization capability. Furthermore, these methods demand a training dataset to train a neural network. However, our method can avoid these restrictions by casting the video prediction problem as optimization. We do not require external training and do not have any assumptions about the data. Our method is very general and outperforms previous external RGB-based methods and methods using additional assumptions~\cite{Wang2018,seg2vid,FVS}. Statistical analysis for long-term prediction and more visual comparisons are provided in the supplement.

\subsection{Evaluation on Driving Datasets}\label{sec1}
We first evaluate our approach and relevant baselines on driving datasets, where semantic information is available, because some baselines require additional semantic maps.

\textbf{Datasets.} Cityscapes ~\cite{Cordts2016Cityscapes} and KITTI datasets~\cite{kitti} contain driving sequences. Our evaluation setting follows ~\cite{FVS}.

\begin{table*}[t]
\vspace{-1mm}
\centering
\resizebox{1\linewidth}{!}{
\begin{tabular}{ccccccccccc}
\toprule
& \multicolumn{4}{c}{DAVIS} & \multicolumn{4}{c}{Middlebury} & \multicolumn{2}{c}{Vimeo90K} \\
\midrule
& \multicolumn{2}{c}{MS-SSIM ($\times1\mathrm{e}{-2}$)$\uparrow$} & 
\multicolumn{2}{c}{LPIPS ($\times1\mathrm{e}{-2}$)$\downarrow$} & 
\multicolumn{2}{c}{MS-SSIM ($\times1\mathrm{e}{-2}$)$\uparrow$} & 
\multicolumn{2}{c}{LPIPS ($\times1\mathrm{e}{-2}$)$\downarrow$} & 
\multicolumn{1}{c}{MS-SSIM ($\times1\mathrm{e}{-2}$)$\uparrow$} & 
\multicolumn{1}{c}{LPIPS ($\times1\mathrm{e}{-2}$)$\downarrow$} \\
& t+1  & t+3 & t+1  & t+3  & t+1  & t+3 & t+1  & t+3 & t+1  & t+1 \\
\midrule
\multicolumn{11}{c}{\textit{External learning methods}} \\
\midrule
DVF~\cite{liu2017voxelflow} & 68.61 & 55.47 & 23.23 & 34.22 & 83.98 & 65.54 & 13.57 & 25.70 & 92.11 & 7.73   \\
DYAN~\cite{dyan} & 78.96 & 70.41 & 13.09 & 21.43 & 92.96 & 83.91 & 7.98 & 15.03 & N/A & N/A          \\
\hline
 \multicolumn{11}{c}{\textit{Optimization methods}} \\
\hline
Ours & \textbf{83.26} & \textbf{73.85} & \textbf{11.40} & \textbf{18.21} & \textbf{94.49} & \textbf{87.96} & \textbf{6.07} & \textbf{10.82} & \textbf{96.75} & \textbf{3.59}   \\
\bottomrule
\end{tabular}
}
\vspace{-1mm}
\caption{Evaluation on diverse datasets. Comparison with state-of-the-art methods on DAVIS, Middlebury, and Vimeo90K.}
\label{table:cross}
\end{table*}

\begin{figure*}[t!]
% \hspace{-6mm}
\centering
\setlength\tabcolsep{0.6mm}
\begin{tabular}{cccc} 
\includegraphics[width=0.24\linewidth]{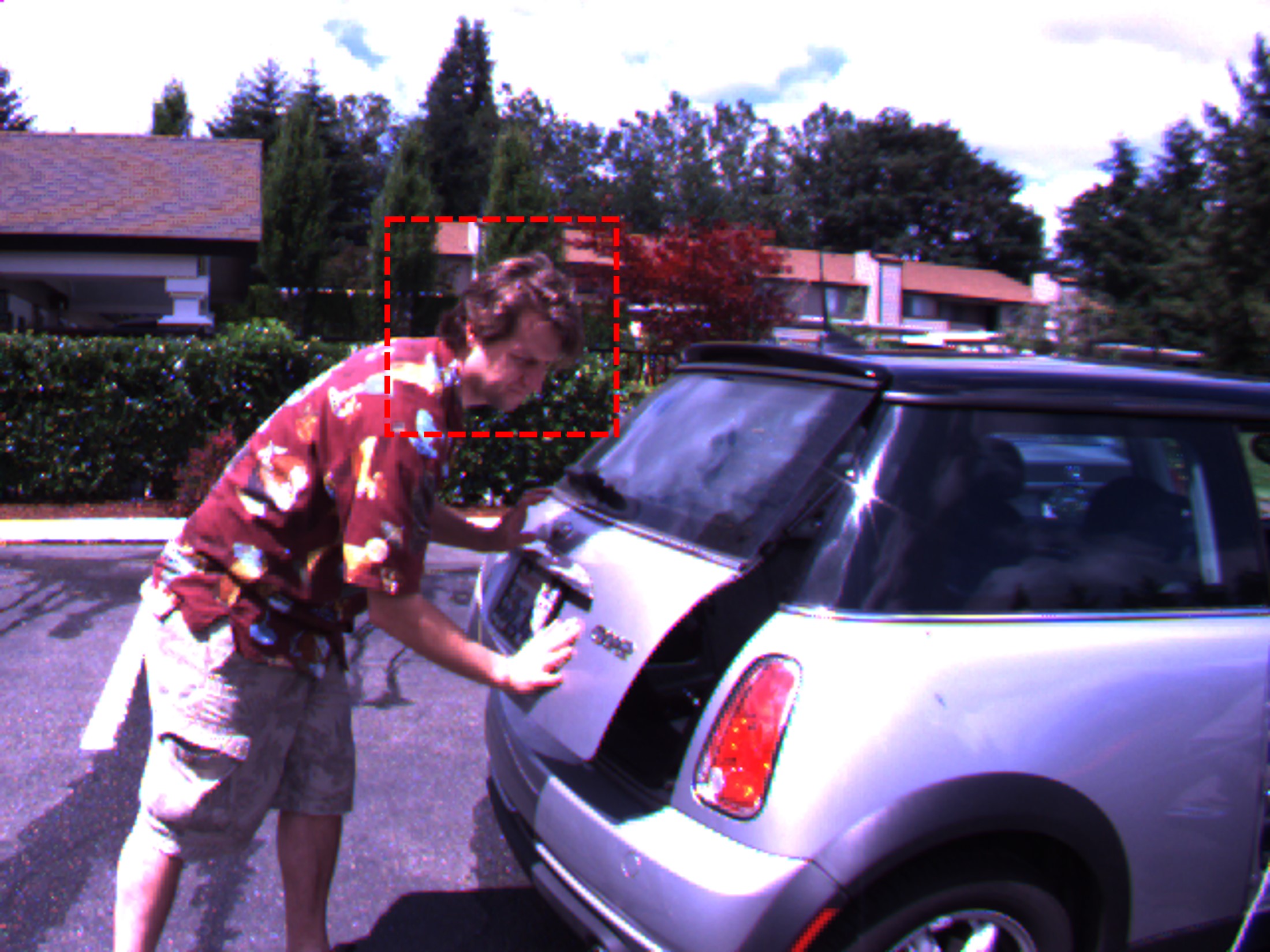}  & 
\includegraphics[width=0.24\linewidth]{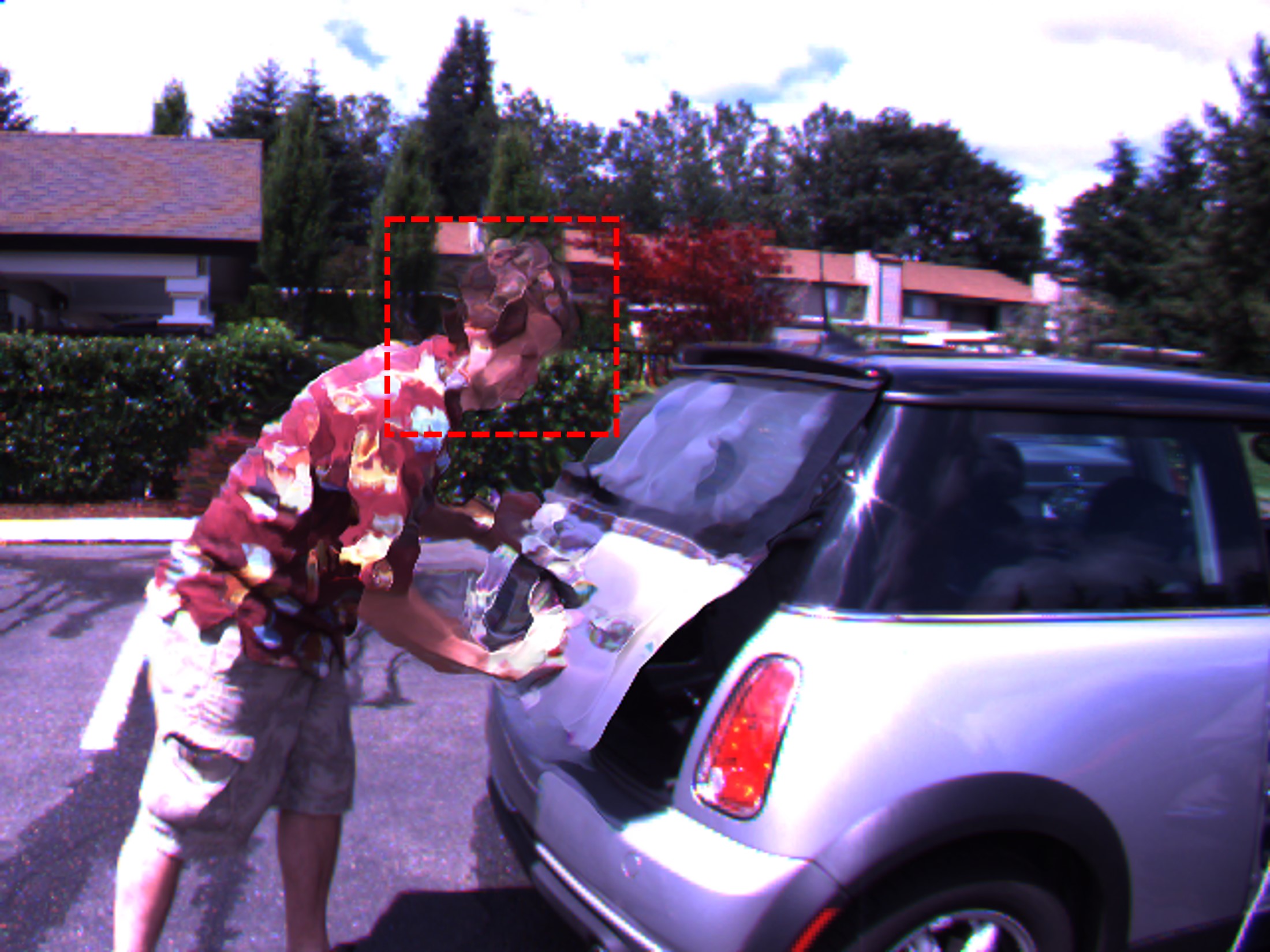} & \includegraphics[width=0.24\linewidth]{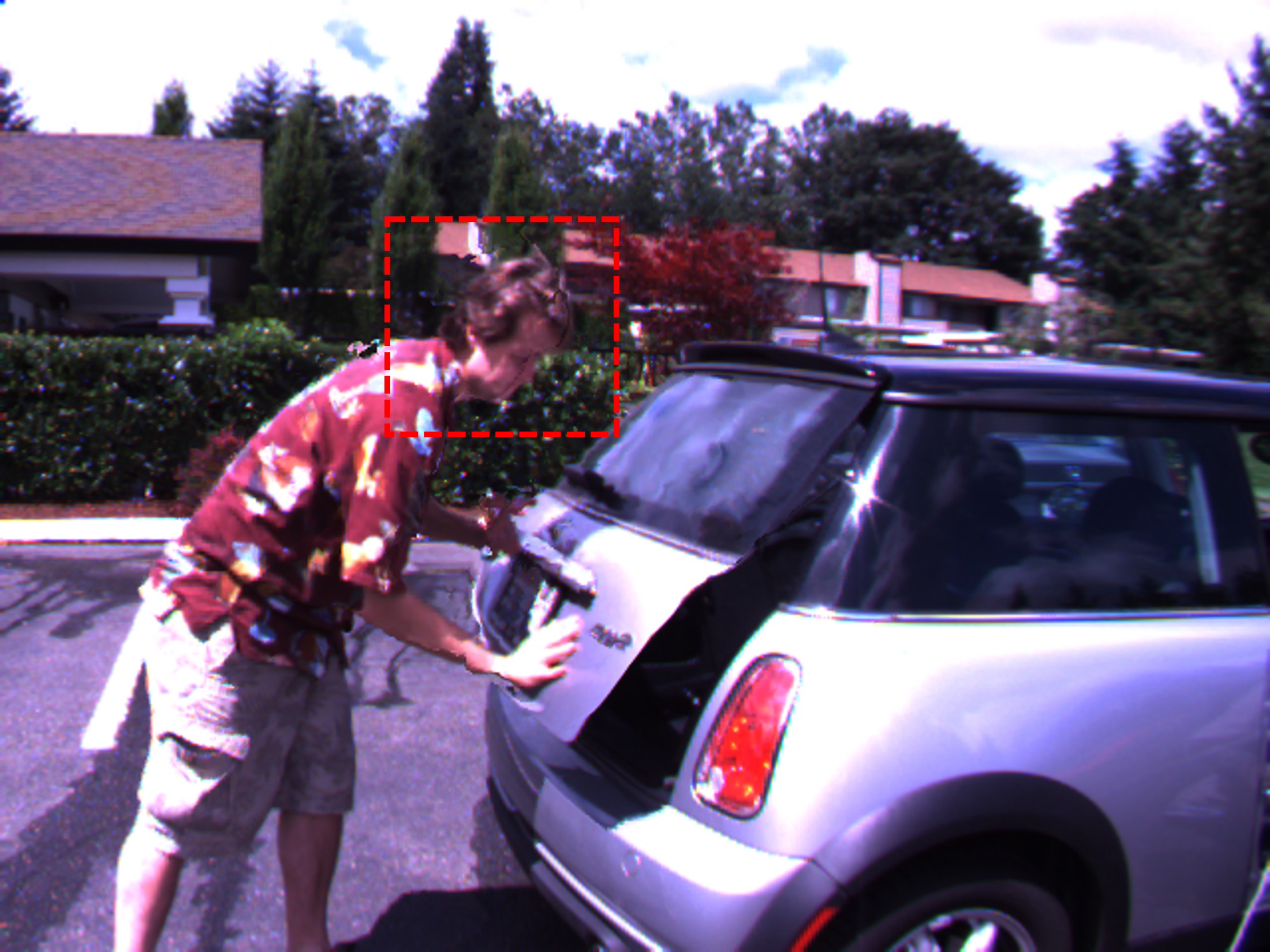} & 
\includegraphics[width=0.24\linewidth]{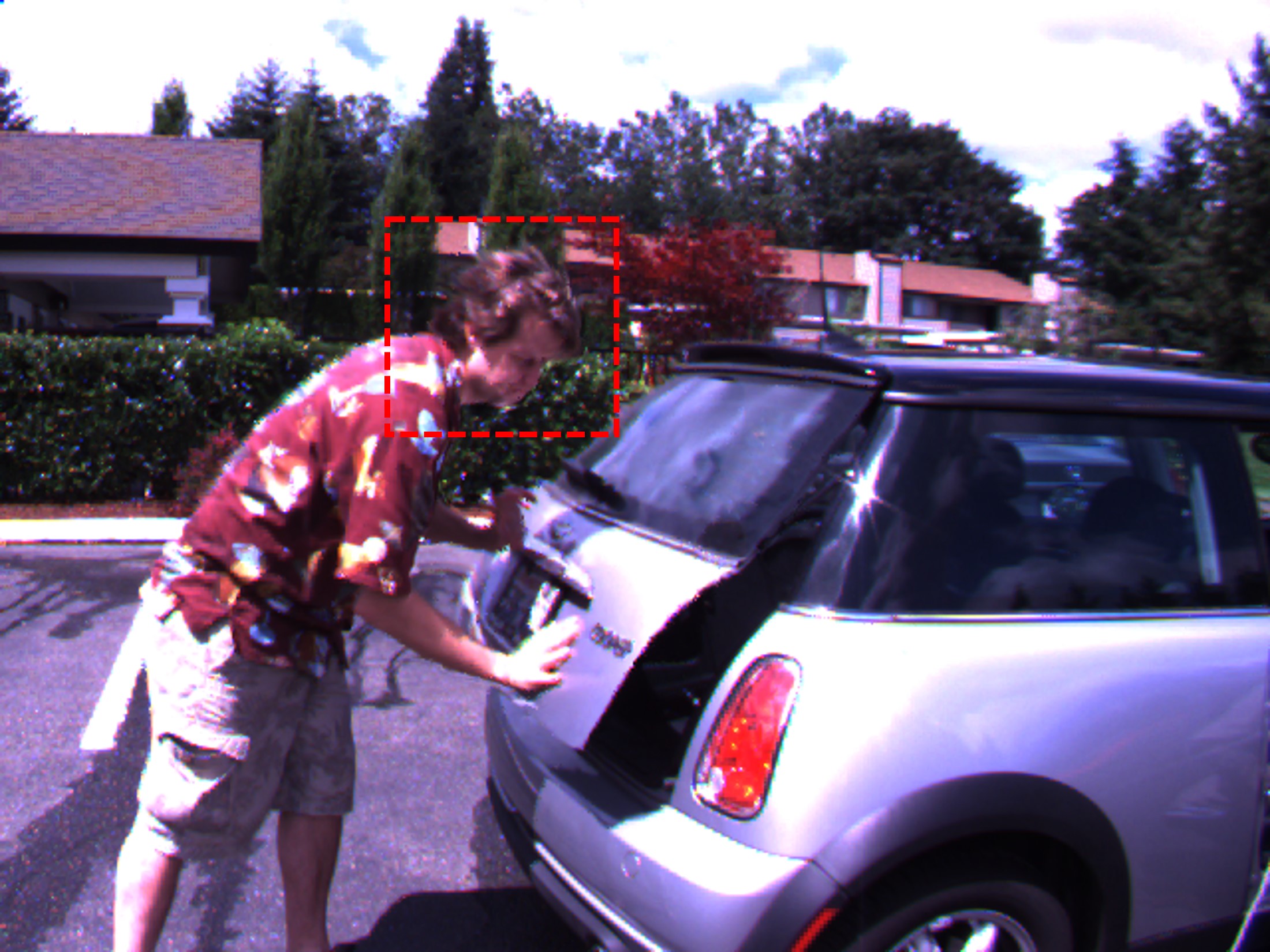} \\ 
\includegraphics[width=0.24\linewidth]{figs/davis_middle/0004_gt_.jpg} &
\includegraphics[width=0.24\linewidth]{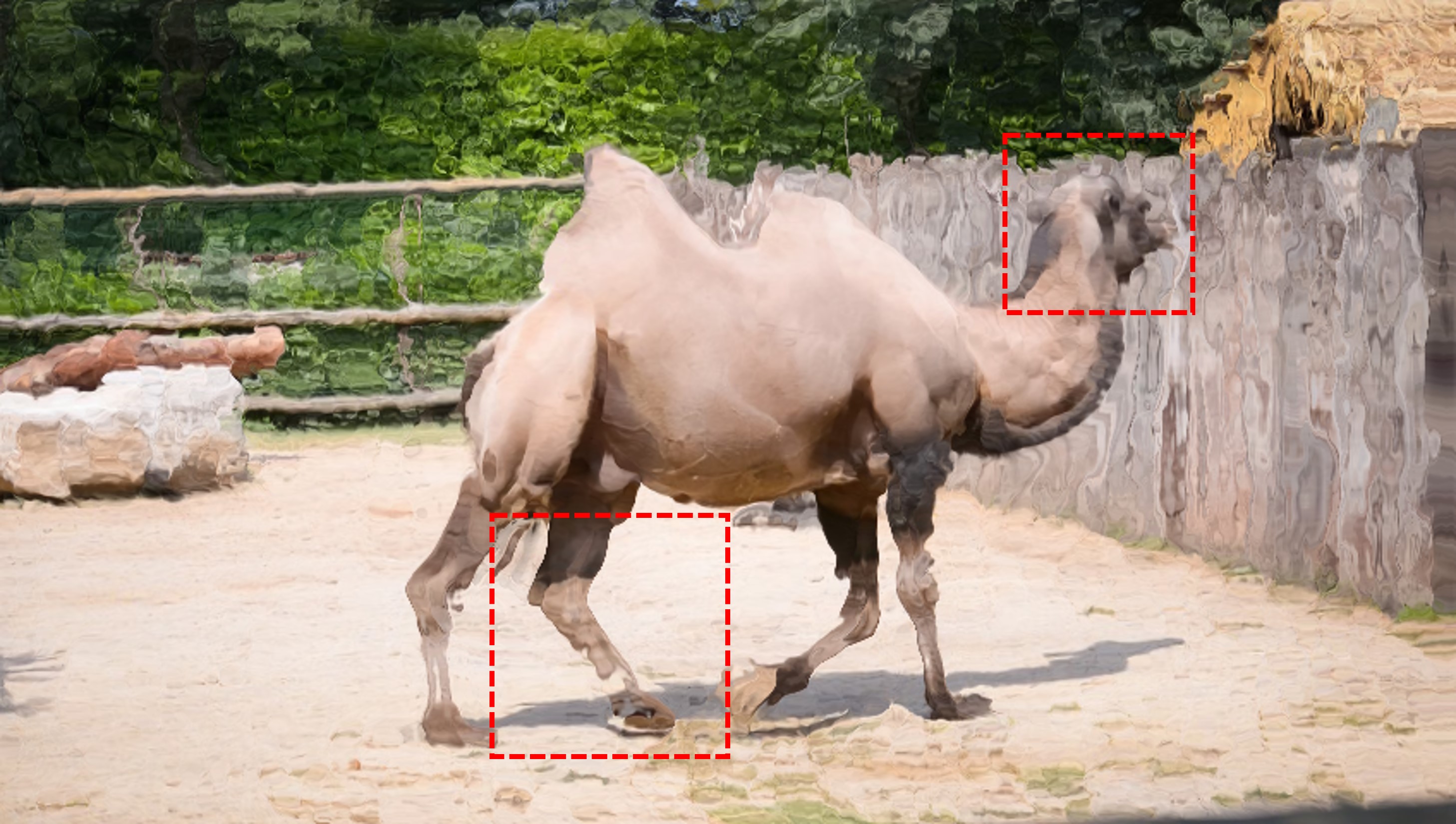}  & 
\includegraphics[width=0.24\linewidth]{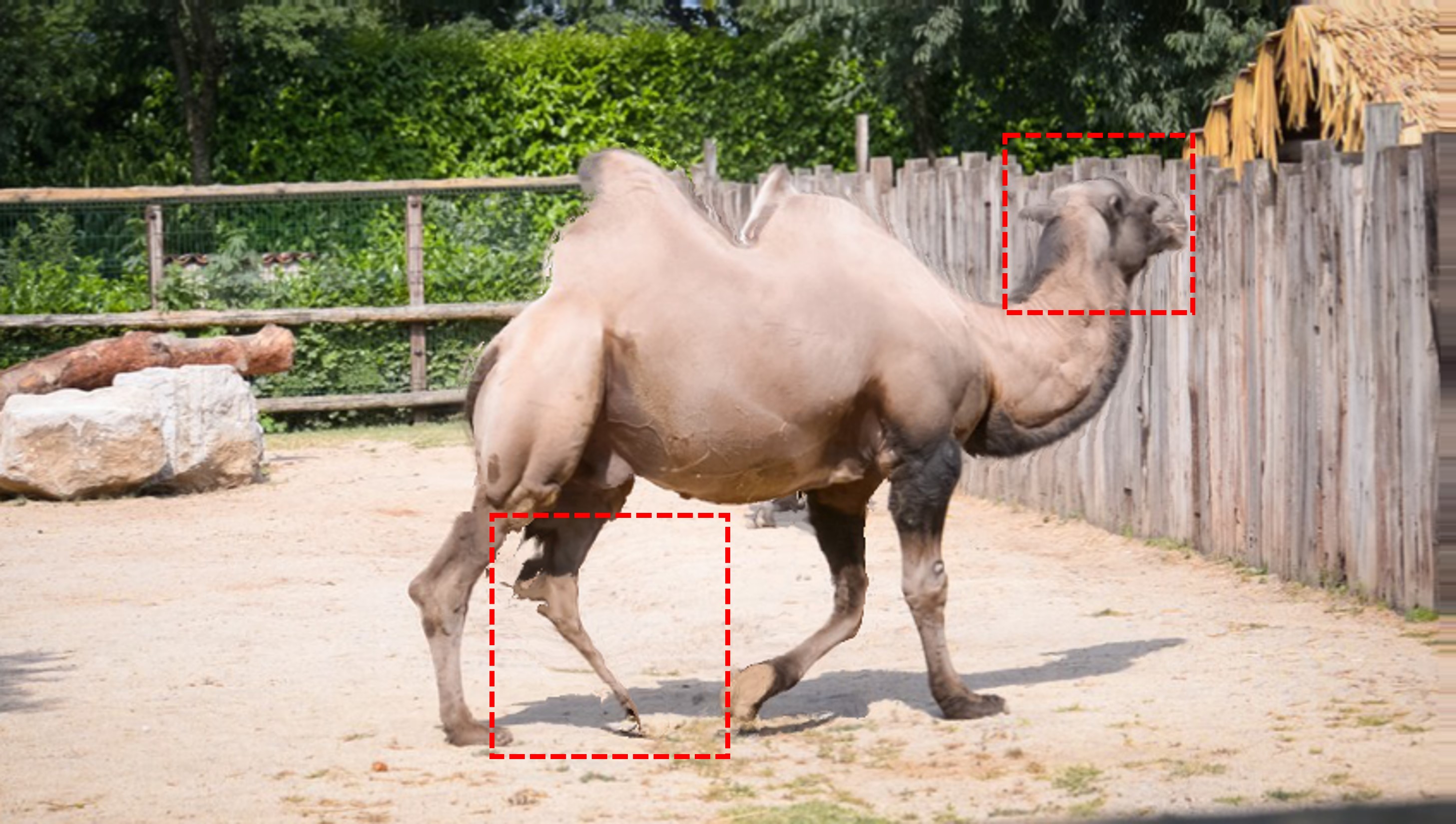} & \includegraphics[width=0.24\linewidth]{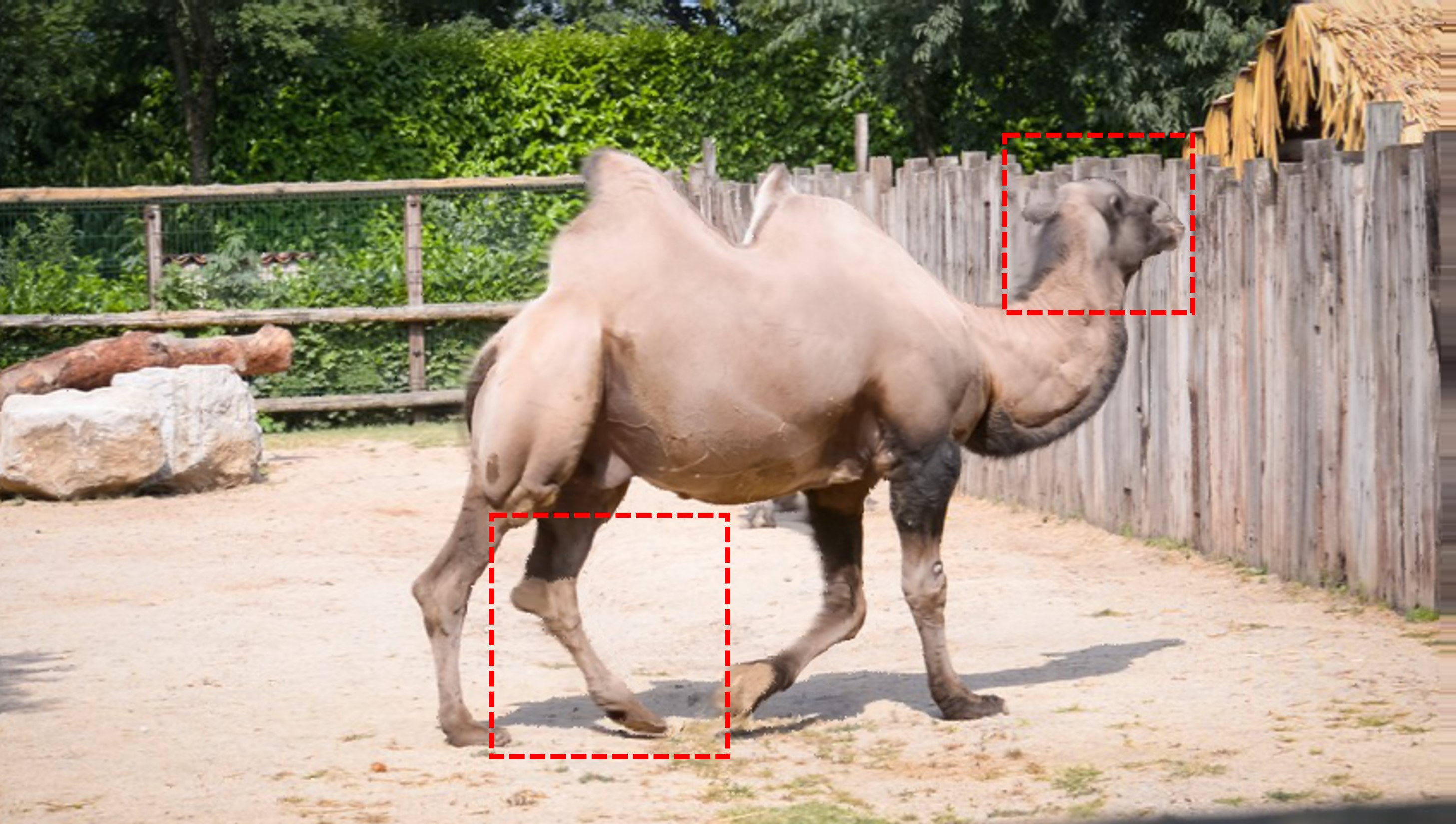} \\ 
\includegraphics[width=0.24\linewidth]{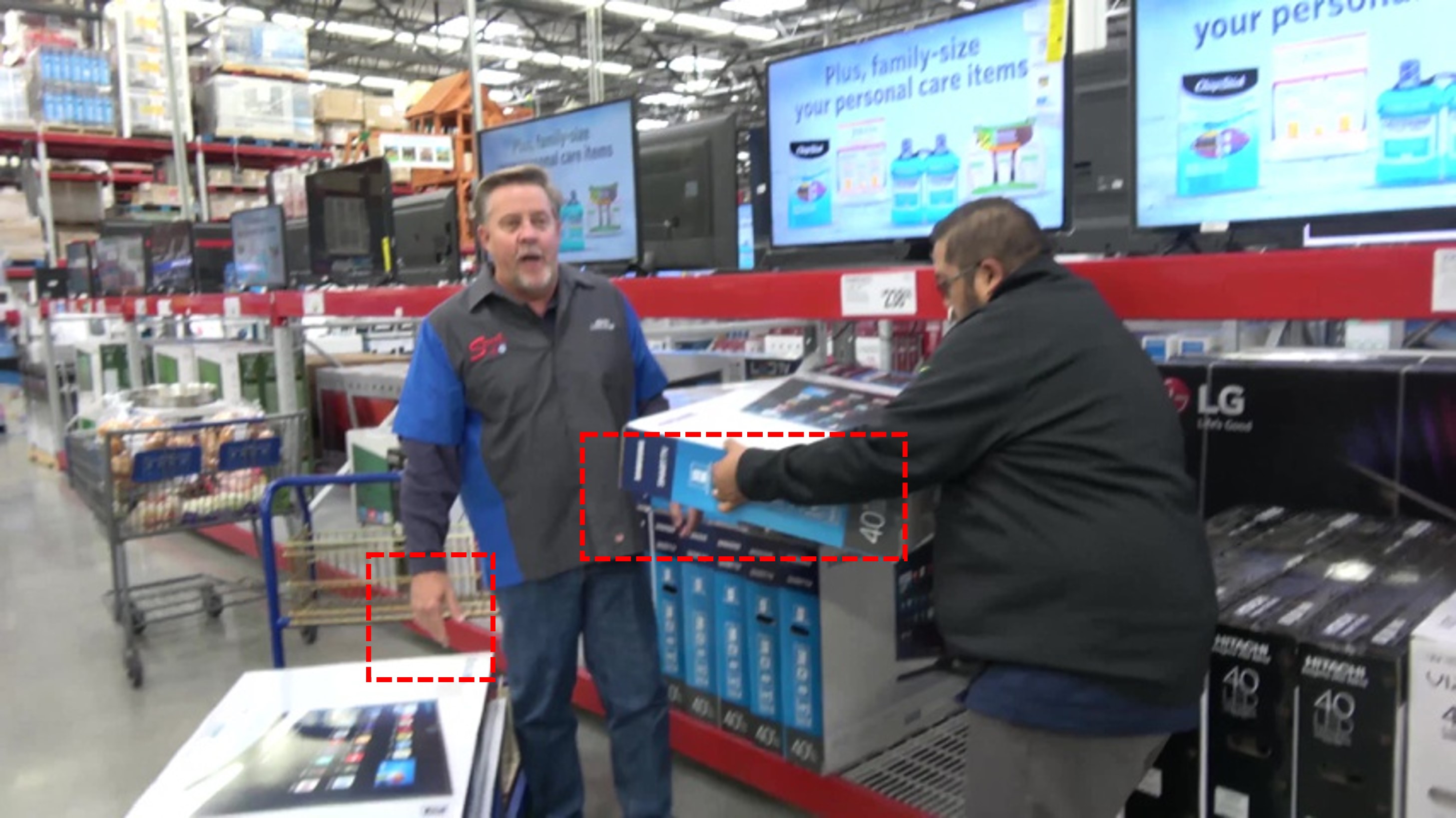}  & 
\includegraphics[width=0.24\linewidth]{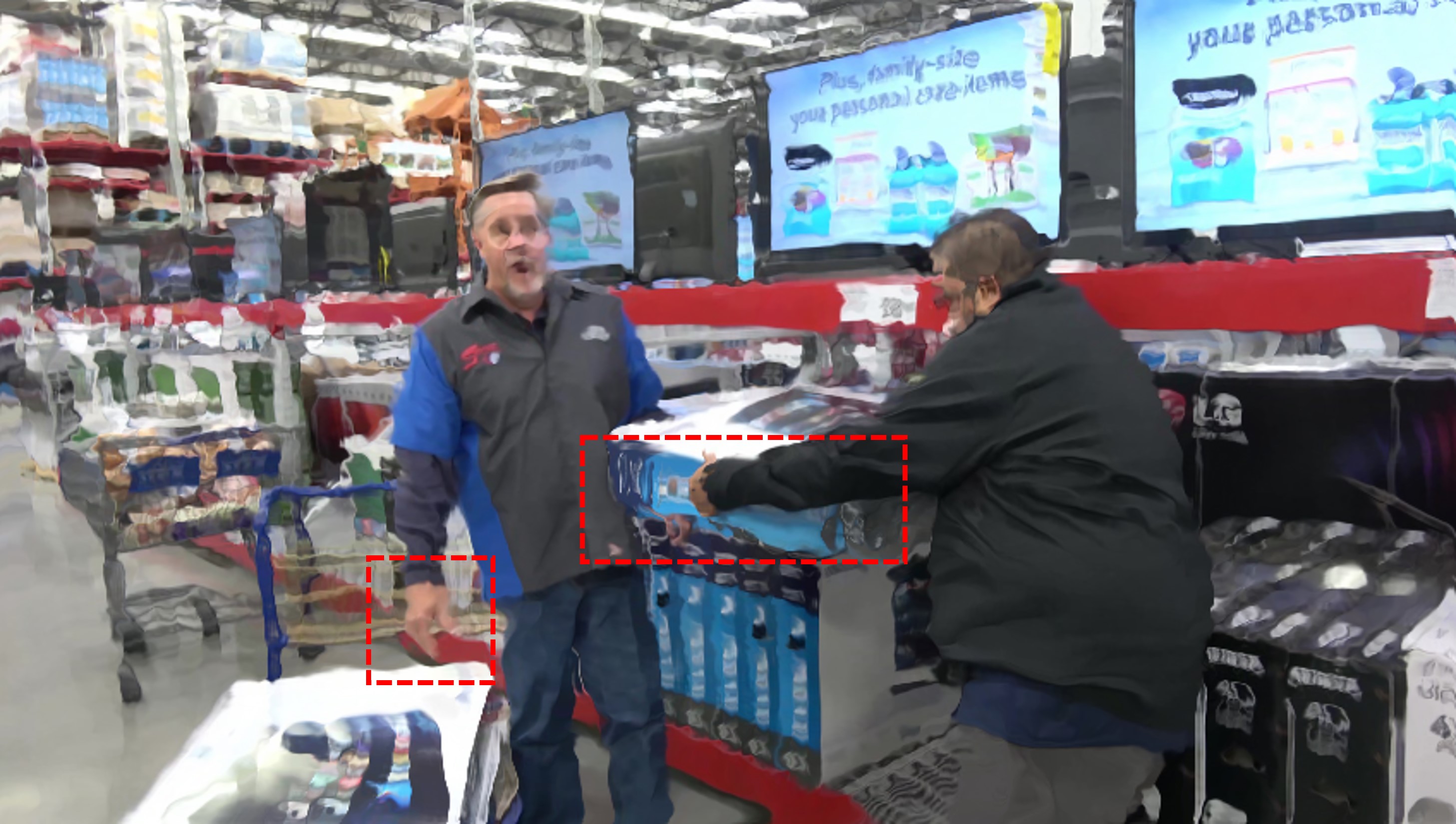} & \includegraphics[width=0.24\linewidth]{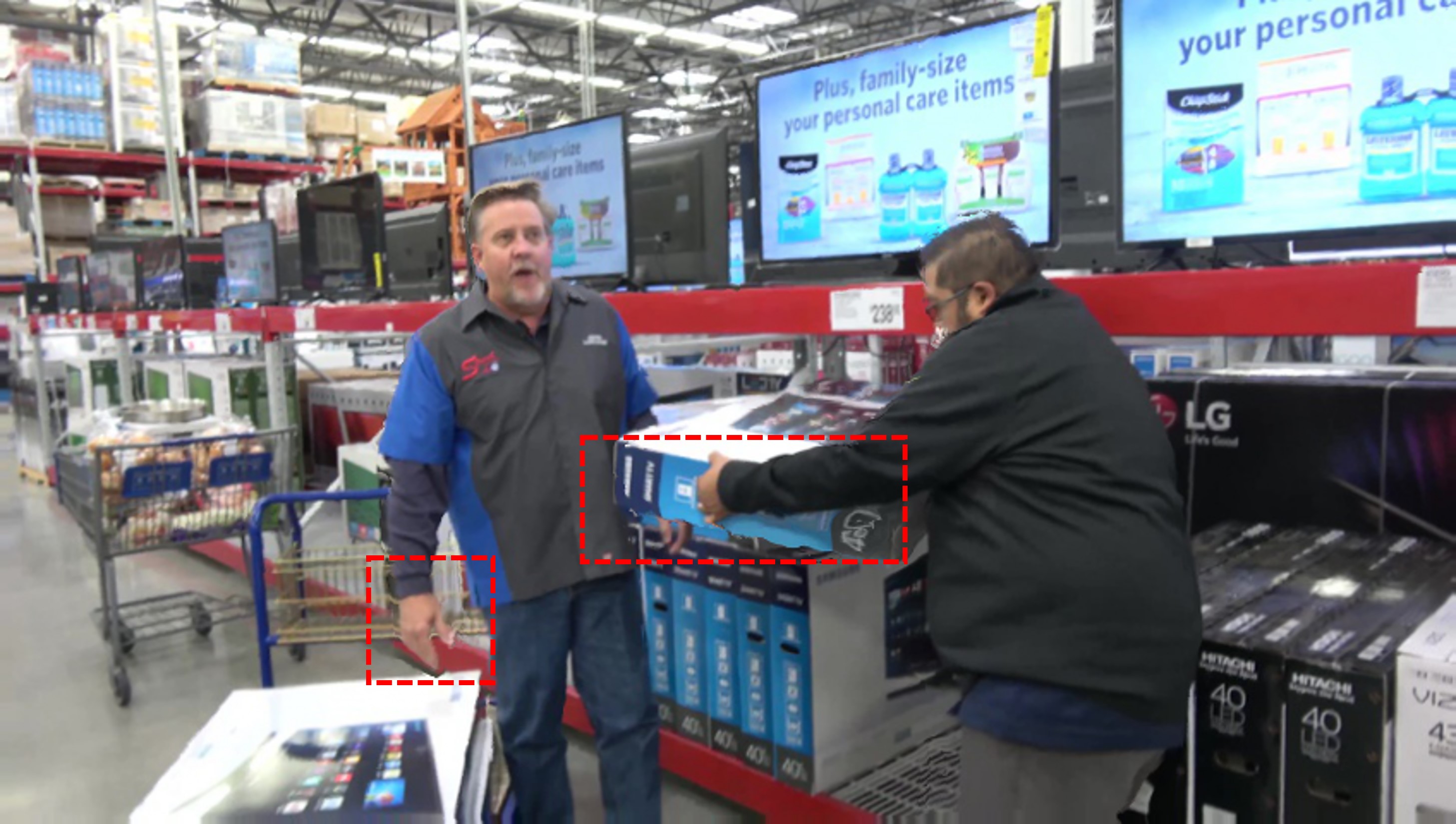} & 
\includegraphics[width=0.24\linewidth]{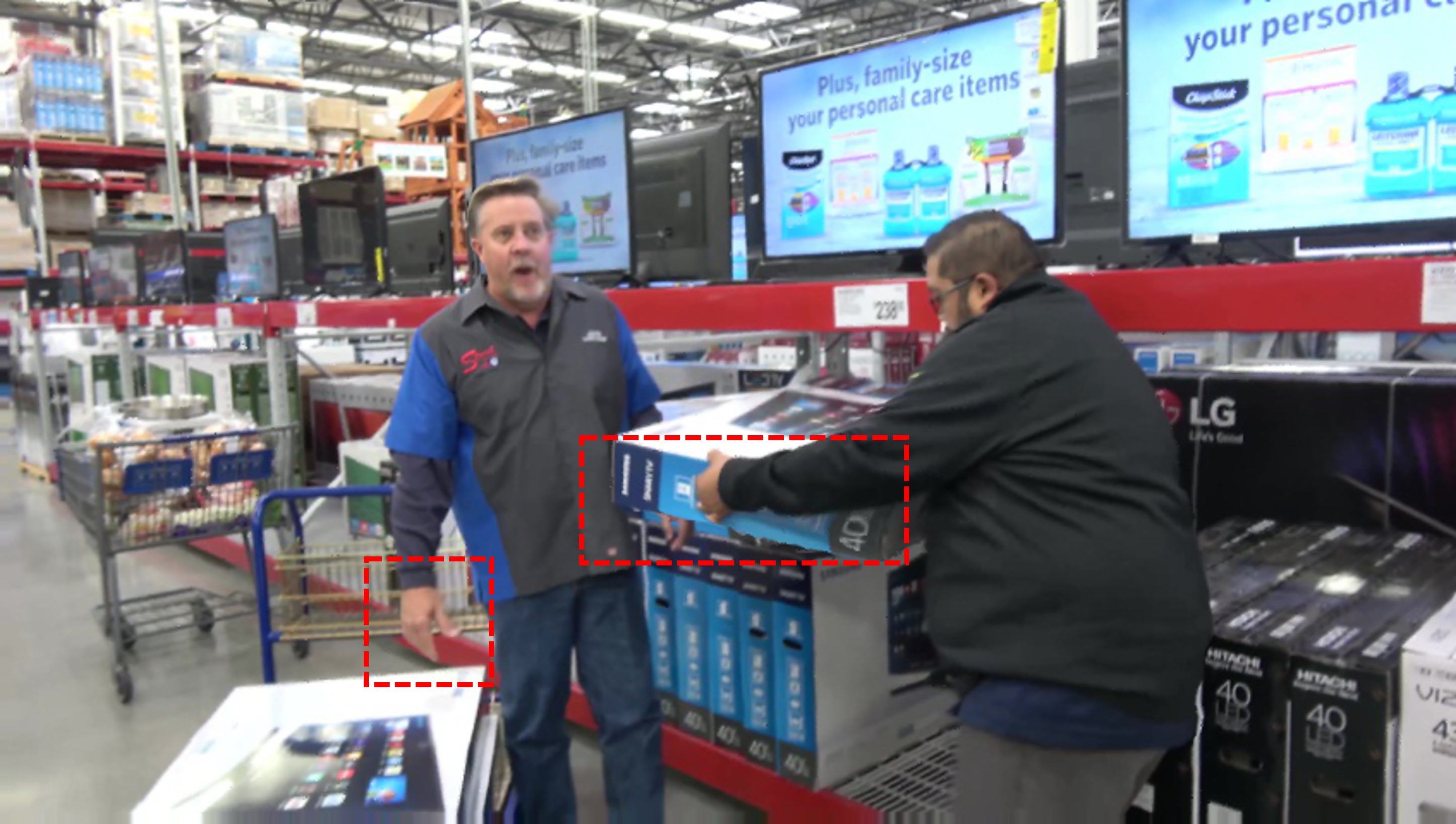} \\ 
GT & DVF~\cite{liu2017voxelflow} & DYAN~\cite{dyan} & Ours
\end{tabular}
\hspace{-4mm}
\caption{Visual comparison on DAVIS and Middlebury datasets. }\label{davis_middlebury}
\hspace{-2mm}
\end{figure*}

\textbf{Baselines.}
All baselines are trained on the corresponding training set in Cityscapes and KITTI.
We categorize baselines into two types. One type is the methods that take only RGB frames as input, such as PredNet~\cite{prednet}, MCNet~\cite{villegas17mcnet}, and DVF~\cite{liu2017voxelflow}. Another type is the methods that require some additional information, such as semantic maps, instance maps, including Vid2vid~\cite{Wang2018}, Seg2vid~\cite{seg2vid}, and FVS~\cite{FVS}. 
%These methods utilize these assumptions to decompose the scene into layers, instances. 
However, these assumptions restrict these methods to be only applicable when this additional context is accessible, degrading their potential generalization ability. We use Multi-scale Structural Similarity Index Measure (MS-SSIM)~\cite{msssim} and LPIPS~\cite{lpips} as evaluation metrics. Higher MS-SSIM and lower LPIPS indicate better performance.

\begin{figure*}[t!]
\vspace{-6mm}
\centering
\setlength\tabcolsep{0.6mm}
\begin{tabular}{ccccc} 
% \rotatebox{90}{\hspace{8mm}$GT_{t+1}$} &
Ground Truth & \multicolumn{2}{c}{DVF~\cite{liu2017voxelflow}} &  \multicolumn{2}{c}{Ours} \\
\includegraphics[ width=0.193\linewidth]{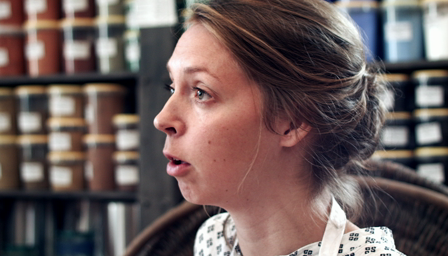}  & 
\includegraphics[ width=0.193\linewidth]{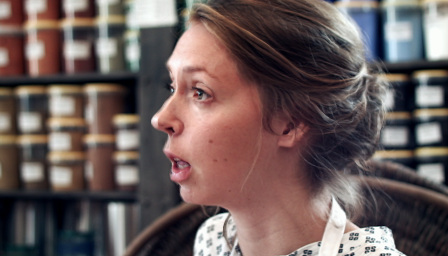} & \includegraphics[ width=0.193\linewidth]{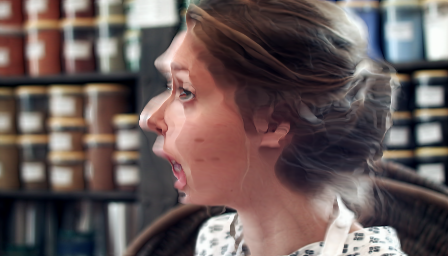} & 
\includegraphics[ width=0.193\linewidth]{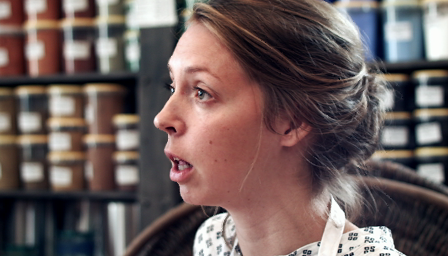} &
\includegraphics[ width=0.193\linewidth]{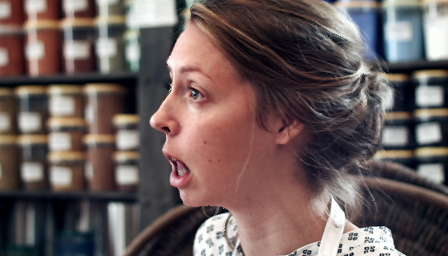}\\ 
% \rotatebox{90}{\hspace{8mm}$t+1$} &
\includegraphics[ width=0.193\linewidth]{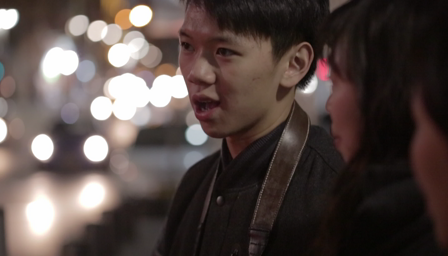}  & 
\includegraphics[ width=0.193\linewidth]{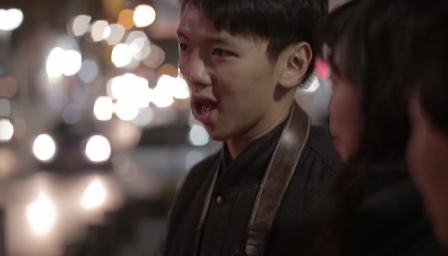} & \includegraphics[ width=0.193\linewidth]{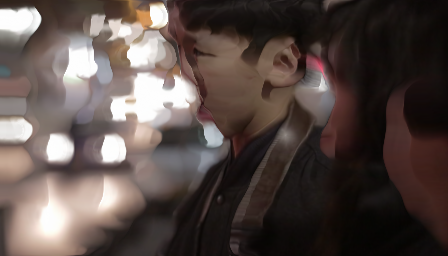} & 
\includegraphics[ width=0.193\linewidth]{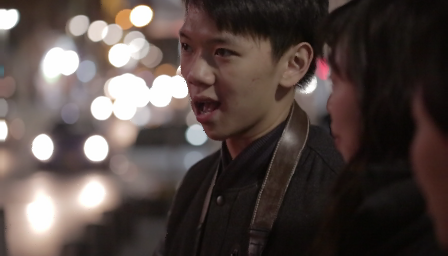} &
\includegraphics[ width=0.193\linewidth]{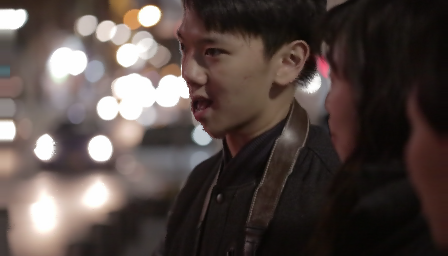} \\ 
% \rotatebox{90}{\hspace{8mm}$t+3$} &
$t+1$ & $t+1$ & $t+3$ & $t+1$ & $t+3$
\end{tabular}
\vspace{-3mm}
\caption{Multi-frame prediction on Vimeo90K. Since Vimeo90K contains triplets and we use two frames as input, there is no ground-truth corresponding to $t+3$.}\label{vimeo}
\vspace{-3mm}
\end{figure*}

\begin{figure*}[t!]
\centering
\includegraphics[width=1\linewidth]{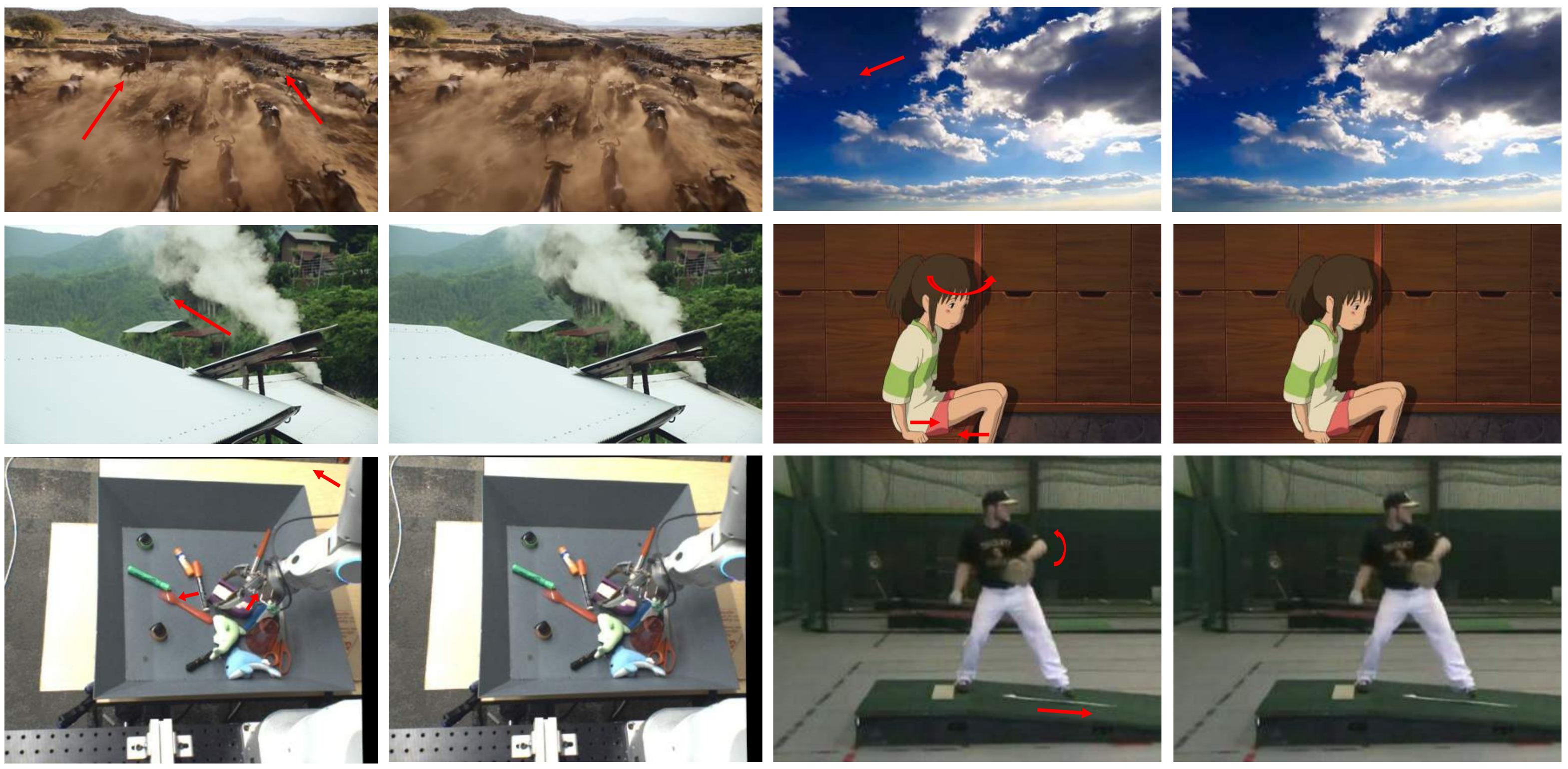}
\hspace{-2mm} Input \hspace{35mm} Ours \hspace{35mm} Input \hspace{36mm} Ours
\vspace{-2mm}
\caption{The next frame prediction performance on diverse data. The left image is the last observed frame, and we use red arrows to indicate the motion from the input frames. The right image is the predicted next frame by our method.}\label{diverse}
\vspace{-3mm}
\end{figure*}

\textbf{Quantitative results.} 
Although our method does not use the training set in Cityscapes and KITTI, our method can still produce outstanding results. As shown in Table~\ref{table:same}, our method can outperform RGB-based methods by a large margin in both short-term and long-term video prediction. Our method improves DVF~\cite{liu2017voxelflow} by 12.75\%, 13.98\%, 13.06\% in  MS-SSIM, 62.81\%, 48.02\%, 38.07\% in of LPIPS on the $t+1$, $t+3$, $t+5$ predictions. Moreover, compared with the methods utilizing semantic or instance segmentation, our method can still outperform FVS~\cite{FVS} by 6.11\%, 7.09\%, 6.24\% in MS-SSIM on Cityscapes, and 33.21\%, 17.55\%, 13.58\% in LPIPS on KITTI, for the $t+1$, $t+3$, $t+5$ predictions.
%The performance of our method is comparable with the latest semantic segmentation based method, SADM~\cite{sadm}. 

The quantitative results demonstrate that our method without external training on the training set of Cityscapes and KITTI, can achieve better performance in both short-term and long-term video prediction. This is because our method is based on the powerful constraints from VFI: recent VFI models can produce superior interpolation results that can be arguably treated as ground truth. While other methods use handcrafted loss functions, these methods may overfit to the major motion exhibited in the training set, the "zooming-in" effect, which is caused by the moving of the forwarding car, instead of learning the real motion.

% And we conduct experiments in the validation subset, consisting of 500 sequences. Our experiment is conducted in $1024 \times 512$ resolution.
%  contains images sequences for driving scenes. We use the test split follows ~\cite{FVS}, and conduct experiments in $832 \times 256$.

\textbf{Qualitative Results.}
In Fig.~\ref{fig:same}, our method is compared to recent video prediction methods, DVF~\cite{liu2017voxelflow} (RGB) and FVS~\cite{FVS} (RGB + semantic + instance). DVF~\cite{liu2017voxelflow} tends to be dominated by ``zooming-in'' motion without predicting the true motion, since the driving datasets are captured by a forward moving camera. FVS~\cite{FVS} uses a handcrafted 2D affine transformation to approximate the motion of moving cars. However, complex motion, including nonrigid deformation and 3D rotation, can not be captured by 2D affine transformation. Moreover, DVF and FVS rely on semantic and instance segmentation, and their performance degrades when these assumptions do not hold. Visual comparisons on Cityscapes are presented in the supplement.
 
\subsection{Evaluation on Diverse Datasets}\label{sec2}
Since our optimization framework does not need external training, our method can be generalized to any video at any resolution. Meanwhile, previous external methods trained on dataset A may have performance degradation when applied to dataset B, caused by the domain gap between A and B. To demonstrate the generality of our method, we conduct a cross-dataset evaluation on diverse datasets. 

\textbf{Datasets.} 
We evaluate our methods on multiple datasets, including DAVIS~\cite{davis}, Middlebury-Other~\cite{middlebury}, and Vimeo90K~\cite{vimeo} datasets.
\textbf{DAVIS~\cite{davis}:} there are 30 sequences in the validation set with resolutions around $854 \times 480$ .
\textbf{Middlebury~\cite{middlebury}:} there are 10 videos with resolutions around $640\times 480$.
\textbf{Vimeo90K~\cite{vimeo}:} there are 3782 triplets in the test set with a resolution of $448 \times 256$. By taking one clip every ten clips, we form the test set.

\begin{table}
%\vspace{-3mm}
\centering
\setlength{\tabcolsep}{3mm}
\resizebox{0.9\linewidth}{!}{
\begin{tabular}{lccc}
\toprule
Components   & SSIM$\uparrow$       & PSNR$\uparrow$       & LPIPS$\downarrow$    \\ 
\midrule
Zero & 0.7719 & 23.79 & 0.1230 \\
Noise        &   0.7669           &   23.64                 &  0.1228                    \\ 
% \textcolor{red}{Copy of $f_{t\rightarrow t-1}$} & 0.9027                      &    29.36                  &  0.0617                    \\ 
\midrule
Without $\mathcal{L}_{img}$   &  0.8939      &  28.86                    & 0.0660                     \\
Without $\mathcal{L}_{cons}$       & 0.8732   & 28.43                     & 0.1221                     \\
$\mathcal{L}_{img}$ with MSE      & 0.8877 & 28.97 & 0.1232 \\
With $\mathcal{L}_{interp}$  & 0.8963 & 28.87 & 0.0693\\
Long-term $\mathcal{L}_{img}$ & 0.6978 & 21.75 & 0.1657\\
\midrule
W/o flow inpainting       &  0.8882     & 28.94                     & 0.1139                     \\ 
\midrule
Full Model     & 0.8975                    & 29.10                     & 0.0646                     \\ 
\bottomrule
\end{tabular}}
\vspace{-2mm}
\caption{The ablation study.}
\vspace{-4mm}
\label{table:ablation}
\end{table}

% \begin{table*}[b]
% \centering
% \resizebox{0.9\linewidth}{!}{
% \begin{tabular}{cccccccccccccccc}
% \toprule
% Iterations & 10 & 100 & 200 & 400  & 600 & 800  & 1000 & 2000 & 3000 & 4000 & 5000 & 6000  & 7000 & 8000 & 9000  \\ 
% \midrule
% LPIPS ($\times1\mathrm{e}{-2}$)\downarrow & 8.31 & 7.08 & 6.98 & 6.89 & 6.86  & 6.84 & 6.80 & 6.76 & 6.73 & 6.73& 6.71 & 6.71 & 6.71 & 6.70 & 6.69\\
% \bottomrule
% \end{tabular}
% }
% \caption{Our model with different optimization iterations in LPIPS.}
% \label{table:convergence}
% \end{table*}

% \begin{table}
% \centering
% \resizebox{0.9\linewidth}{!}{
% \begin{tabular}{ccccccc}
% \toprule
% Iterations & 10 & 100 & 400 & 1000 & 3000 & 9000  \\ 
% \midrule
% LPIPS ($\times1\mathrm{e}{-2}$)\downarrow & 8.31 & 7.08 & 6.89 & 6.80 & 6.73 &6.69\\
% \bottomrule
% \end{tabular}
% }
% \caption{Our model with different optimization iterations.}
% \vspace{-2mm}
% \label{table:convergence}
% \end{table}

\textbf{Baselines.}
We compare our method with two latest external methods, DVF~\cite{liu2017voxelflow} and DYAN~\cite{dyan}, which take only RGB frames as input, since they can be applied to these datasets. Other methods that require additional assumptions cannot be compared because their assumptions do not hold. We test these two models on these datasets using their pretrained model on UCF101~\cite{ucf101}.

\textbf{Quantitative results.}
As exhibited in Table~\ref{table:cross}, our method outperforms DYAN~\cite{dyan} by 12.90\%, 15.0\% in DAVIS, and by 23.97\%, 28.00\% on Middlebury in LPIPS for $t+1$, $t+3$ predictions. Note that our method is still robust in long-term prediction. On the Vimeo90K dataset, our method outperforms DVF~\cite{liu2017voxelflow} by 53.56\% in LPIPS for next-frame prediction. Although these two baselines may perform well on UCF-101, there is a domain gap problem between UCF-101 and these test videos. This domain gap phenomenon commonly exists in video prediction tasks, so most video prediction methods~\cite{prednet, villegas17mcnet, liu2017voxelflow, dyan, seg2vid, FVS, sadm} usually train a separate model for each dataset and even employ different assumptions for each dataset. Differently, the domain gap in the VFI task does not appear to be an issue. VFI methods~\cite{RIFE,park2021ABME} can use one dataset for training and can produce excellent results on various datasets. With the powerful constraints provided by VFI, our method does not have the domain gap problem (no external training): each sequence is independently optimized by employing the FVI network as a constraint. 

\textbf{Qualitative results.}
As shown in Fig.~\ref{davis_middlebury} and Fig.~\ref{vimeo}, our method yields better prediction results than baselines. The baselines produce distortion artifacts around object boundary or fail at prediction when motion is complicated. Meanwhile, our method can robustly predict future frames. As the Vimeo90K dataset only provides three frames, we take the first two frames as input for future frame prediction (there is only ground truth for the $t+1$ prediction). Our method can produce high-quality prediction results in the long term, as shown in Fig.~\ref{vimeo}. The intricate details, such as the hair of the woman, remain clear in our results.% Moreover, the recent popular semantic segmentation-based video prediction frameworks cannot achieve these performances.

To demonstrate the generality of our method, we collect some real videos from YouTube, such as the movie clips as shown in Fig.~\ref{diverse}. We also present some visual results on the BAIR Pushing ~\cite{bair_pushing} and Penn Action~\cite{penn_action} datasets. Our method can produce strong results on diverse data.
%Our method can be applied to any video at any resolution theoretically.

\subsection{Ablation Study}\label{sec3}
We conduct an ablation study to demonstrate the importance of each component, as shown in Table~\ref{table:ablation}.
Our ablation study is conducted on Cityscapes. We use the performance of the next-frame prediction to evaluate different components. Visual comparisons are presented in the supplement.

\textbf{Initialization.} If we set the optimization target as the prediction frame, then the optimization cannot converge. Thus, we choose to optimize optical flow, which eases the optimization process since pixels can be copied from observed frames. If we initialize optical flow as zeros or Gaussian noise, then our performance becomes worse. If we initialize the optical flow as the copy of $f_{t\rightarrow t-1}$, then the performance is similar to our full model. 
This demonstrates that good initialization helps the optimization to converge.
%This demonstrates that good initialization helps the optimization towards the correct optimization direction.

\textbf{Loss functions.} We conduct experiments with loss functions with several variants.
Without $\mathcal{L}_{img}$: supervised only by $\mathcal{L}_{cons}$.
Without $\mathcal{L}_{cons}$: supervised only by $\mathcal{L}_{img}$. 
$\mathcal{L}_{img}$ with MSE: using MSE loss rather than $L_1$ loss in Eq.~\ref{eq:img}. 
With $\mathcal{L}_{interp}$: use the interpolation result of VFI $I^G_{t}$ rather than the intermediate output $I^G_{t+1}$ in Eq.~\ref{eq:img}. 
Long-term $\mathcal{L}_{img}$: setting the input frame length as 4 and add a long-term constraint between $x_{t-3}$, $\tilde{x}_{t+1}$ and $x_{t-1}$.
%besides constrains between $x_{t-1}$, $x_{t}$, $\tilde{x}_{t+1}$.
The results show that the combination of $\mathcal{L}_{img}$ and $\mathcal{L}_{cons}$ performs best. $\mathcal{L}_{img}$ provides a more direct constraint than $\mathcal{L}_{interp}$. Long-term $\mathcal{L}_{img}$ has performance degradation because when the motion is too large, the accuracy of VFI decreases .

\textbf{Flow inpainting.} If we remove the optical flow inpainting procedure, then our performance also drops because optical flow inpainting effectively corrects invalid flow values.

\textbf{Other pretrained VFI models.}
We also try utilizing Super SloMo~\cite{Super_SloMo} as our VFI backbone and find this method also works for our framework. For the next frame prediction on Cityscapes, the MS-SSIM for Super SloMo~\cite{Super_SloMo} and RIFE~\cite{RIFE} is 0.9199 and 0.9454. Thus, we choose RIFE~\cite{RIFE} as our VFI backbone.

\begin{figure}[t!]
\centering
\includegraphics[width=\linewidth]{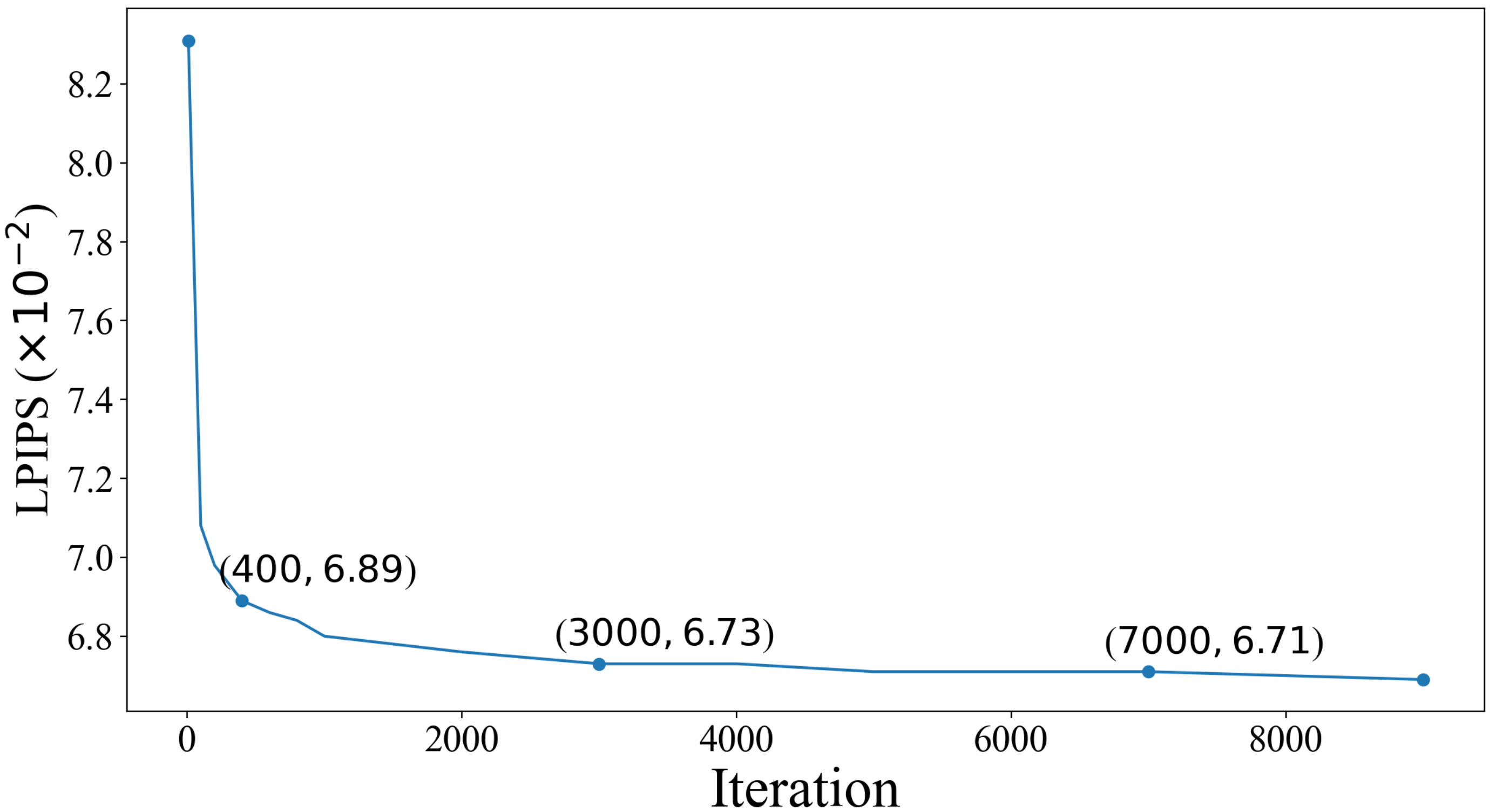}
\vspace{-5mm}
\caption{Our model with different optimization iterations.}\label{converge}
\vspace{-4mm}
\end{figure}

\subsection{Convergence Analysis}
As shown in Fig.~\ref{converge}, we conduct the convergence analysis on Cityscapes with a resolution of 256 $\times$ 512 . In the first 400 iterations, the optimization converges fast. After 400 iterations, the prediction result gradually improves. The prediction result is still slowly improving after 3000 iterations.

\section{Conclusion}
%Given the severe domain gap problem existing in current external video prediction methods, 
We propose the first video prediction optimization method by casting the video prediction problem as a VFI based optimization problem, which addresses the domain gap issue in most video prediction methods. Our method can outperform state-of-the-art methods and can be adapted to any video at any resolution.
Although our method relieves domain gap problem and presents impressive performance, optimizing every frame by our method costs more time than other external learning-based methods. The comparison between our and other methods in terms of the model size and inference time is presented in the supplement. As we observe, most run time in our model is spent on the gradient propagation inside the VFI network \cite{RIFE}, which inspires us to design a more efficient backbone for acceleration in the future.

\newpage
{\small
\bibliographystyle{ieee_fullname}
\bibliography{egbib}

\begin{thebibliography}{10}\itemsep=-1pt

\bibitem{middlebury}
Simon Baker, Daniel Scharstein, J.~P. Lewis, Stefan Roth, Michael~J. Black, and
  Richard Szeliski.
\newblock A database and evaluation methodology for optical flow.
\newblock In {\em ICCV}, 2007.

\bibitem{bao2019dain}
Wenbo Bao, Wei-Sheng Lai, Chao Ma, Xiaoyun Zhang, Zhiyong Gao, and Ming-Hsuan
  Yang.
\newblock Depth-aware video frame interpolation.
\newblock In {\em CVPR}, 2019.

\bibitem{bao2018memc}
Wenbo Bao, Wei-Sheng Lai, Xiaoyun Zhang, Zhiyong Gao, and Ming-Hsuan Yang.
\newblock {MEMC-Net}: Motion estimation and motion compensation driven neural
  network for video interpolation and enhancement.
\newblock {\em TPAMI}, 43(3):933--948, Mar. 2021.

\bibitem{sadm}
Xinzhu Bei, Yanchao Yang, and Stefano Soatto.
\newblock Learning semantic-aware dynamics for video prediction.
\newblock In {\em CVPR}, 2021.

\bibitem{Cordts2016Cityscapes}
Marius Cordts, Mohamed Omran, Sebastian Ramos, Timo Rehfeld, Markus Enzweiler,
  Rodrigo Benenson, Uwe Franke, Stefan Roth, and Bernt Schiele.
\newblock The cityscapes dataset for semantic urban scene understanding.
\newblock In {\em CVPR}, 2016.

\bibitem{bair_pushing}
Chelsea Finn, Ian Goodfellow, and Sergey Levine.
\newblock Unsupervised learning for physical interaction through video
  prediction.
\newblock In {\em NeurIPS}, 2016.

\bibitem{DPG}
Hang Gao, Huazhe Xu, Qi-Zhi Cai, Ruth Wang, Fisher Yu, and Trevor Darrell.
\newblock Disentangling propagation and generation for video prediction.
\newblock In {\em ICCV}, 2019.

\bibitem{gatys2016image}
Leon~A Gatys, Alexander~S Ecker, and Matthias Bethge.
\newblock Image style transfer using convolutional neural networks.
\newblock In {\em CVPR}, 2016.

\bibitem{kitti}
Andreas Geiger, Philip Lenz, Christoph Stiller, and Raquel Urtasun.
\newblock Vision meets robotics: The kitti dataset.
\newblock {\em I. J. Robotics Res.}, 2013.

\bibitem{gui2020feflow}
Shurui Gui, Chaoyue Wang, Qihua Chen, and Dacheng Tao.
\newblock {FeatureFlow}: Robust video interpolation via structure-to-texture
  generation.
\newblock In {\em CVPR}, 2020.

\bibitem{RIFE}
Zhewei Huang, Tianyuan Zhang, Wen Heng, Boxin Shi, and Shuchang Zhou.
\newblock {RIFE:} real-time intermediate flow estimation for video frame
  interpolation.
\newblock {\em arXiv preprint arXiv:2011.06294}, 2020.

\bibitem{jaderberg2015spatial}
Max Jaderberg, Karen Simonyan, Andrew Zisserman, et~al.
\newblock Spatial transformer networks.
\newblock {\em NIPS}, 2015.

\bibitem{Super_SloMo}
Huaizu Jiang, Deqing Sun, Varun Jampani, Ming{-}Hsuan Yang, Erik~G.
  Learned{-}Miller, and Jan Kautz.
\newblock Super slomo: High quality estimation of multiple intermediate frames
  for video interpolation.
\newblock In {\em CVPR}, 2018.

\bibitem{kingma2014adam}
Diederik~P Kingma and Jimmy Ba.
\newblock Adam: A method for stochastic optimization.
\newblock In {\em ICLR}, 2015.

\bibitem{revisit21}
Wonkwang Lee, Whie Jung, Han Zhang, Ting Chen, Jing~Yu Koh, Thomas~E. Huang,
  Hyungsuk Yoon, Honglak Lee, and Seunghoon Hong.
\newblock Revisiting hierarchical approach for persistent long-term video
  prediction.
\newblock In {\em ICLR}, 2021.

\bibitem{lei2020dvp}
Chenyang Lei, Yazhou Xing, and Qifeng Chen.
\newblock Blind video temporal consistency via deep video prior.
\newblock In {\em NeurIPS}, 2020.

\bibitem{DVP_lei}
Chenyang Lei, Yazhou Xing, Hao Ouyang, and Qifeng Chen.
\newblock Deep video prior for video consistency and propagation.
\newblock {\em TPAMI}, To Appear.

\bibitem{dyan}
Wenqian Liu, Abhishek Sharma, Octavia Camps, and Mario Sznaier.
\newblock Dyan: A dynamical atoms-based network for video prediction.
\newblock In {\em ECCV}, 2018.

\bibitem{liu2019cyclicgen}
Yu-Lun Liu, Yi-Tung Liao, Yen-Yu Lin, and Yung-Yu Chuang.
\newblock Deep video frame interpolation using cyclic frame generation.
\newblock In {\em AAAI}, 2019.

\bibitem{liu2017voxelflow}
Ziwei Liu, Raymond Yeh, Yiming~Liu Xiaoou~Tang, , and Aseem Agarwala.
\newblock Video frame synthesis using deep voxel flow.
\newblock In {\em ICCV}, 2017.

\bibitem{liu2017dvf}
Ziwei Liu, Raymond~A Yeh, Xiaoou Tang, Yiming Liu, and Aseem Agarwala.
\newblock Video frame synthesis using deep voxel flow.
\newblock In {\em ICCV}, 2017.

\bibitem{prednet}
William Lotter, Gabriel Kreiman, and David Cox.
\newblock Deep predictive coding networks for video prediction and unsupervised
  learning.
\newblock In {\em ICLR}, 2017.

\bibitem{NextSegmPredICCV17}
Pauline Luc, Natalia Neverova, Camille Couprie, Jacob Verbeek, and Yann LeCun.
\newblock Predicting deeper into the future of semantic segmentation.
\newblock In {\em ICCV}, 2017.

\bibitem{meyer2018phase}
Simone Meyer, Abdelaziz Djelouah, Brian McWilliams, Alexander Sorkine-Hornung,
  Markus Gross, and Christopher Schroers.
\newblock {PhaseNet} for video frame interpolation.
\newblock In {\em CVPR}, 2018.

\bibitem{meyer2015phase}
Simone Meyer, Oliver Wang, Henning Zimmer, Max Grosse, and Alexander
  Sorkine-Hornung.
\newblock Phase-based frame interpolation for video.
\newblock In {\em CVPR}, 2015.

\bibitem{mildenhall2020nerf}
Ben Mildenhall, Pratul~P Srinivasan, Matthew Tancik, Jonathan~T Barron, Ravi
  Ramamoorthi, and Ren Ng.
\newblock Nerf: Representing scenes as neural radiance fields for view
  synthesis.
\newblock In {\em European conference on computer vision}, pages 405--421.
  Springer, 2020.

\bibitem{niklaus2018ctx}
Simon Niklaus and Feng Liu.
\newblock Context-aware synthesis for video frame interpolation.
\newblock In {\em CVPR}, 2018.

\bibitem{niklaus2020softsplatting}
Simon Niklaus and Feng Liu.
\newblock Softmax splatting for video frame interpolation.
\newblock In {\em CVPR}, 2020.

\bibitem{niklaus2017adaconv}
Simon Niklaus, Long Mai, and Feng Liu.
\newblock Video frame interpolation via adaptive convolution.
\newblock In {\em CVPR}, 2017.

\bibitem{niklaus2017sepconv}
Simon Niklaus, Long Mai, and Feng Liu.
\newblock Video frame interpolation via adaptive separable convolution.
\newblock In {\em ICCV}, 2017.

\bibitem{seg2vid}
Junting Pan, Chengyu Wang, Xu Jia, Jing Shao, Lu Sheng, Junjie Yan, and
  Xiaogang Wang.
\newblock Video generation from single semantic label map.
\newblock In {\em CVPR}, 2019.

\bibitem{park2020bmbc}
Junheum Park, Keunsoo Ko, Chul Lee, and Chang-Su Kim.
\newblock {BMBC}: Bilateral motion estimation with bilateral cost volume for
  video interpolation.
\newblock In {\em ECCV}, 2020.

\bibitem{park2021ABME}
Junheum Park, Chul Lee, and Chang-Su Kim.
\newblock Asymmetric bilateral motion estimation for video frame interpolation.
\newblock In {\em ICCV}, 2021.

\bibitem{davis}
Jordi Pont{-}Tuset, Federico Perazzi, Sergi Caelles, Pablo Arbelaez, Alexander
  Sorkine{-}Hornung, and Luc~Van Gool.
\newblock The 2017 {DAVIS} challenge on video object segmentation.
\newblock {\em CoRR}, abs/1704.00675, 2017.

\bibitem{Qi2019}
Xiaojuan Qi, Zhengzhe Liu, Qifeng Chen, and Jiaya Jia.
\newblock 3d motion decomposition for {RGBD} future dynamic scene synthesis.
\newblock In {\em CVPR}, 2019.

\bibitem{shaham2019singan}
Tamar~Rott Shaham, Tali Dekel, and Tomer Michaeli.
\newblock Singan: Learning a generative model from a single natural image.
\newblock In {\em ICCV}, 2019.

\bibitem{XVFI}
Hyeonjun Sim, Jihyong Oh, and Munchurl Kim.
\newblock {XVFI}: Extreme video frame interpolation.
\newblock In {\em ICCV}, 2021.

\bibitem{ucf101}
Khurram Soomro, Amir~Roshan Zamir, and Mubarak Shah.
\newblock {UCF101:} {A} dataset of 101 human actions classes from videos in the
  wild.
\newblock {\em CoRR}, abs/1212.0402, 2012.

\bibitem{Srivastava2015}
Nitish Srivastava, Elman Mansimov, and Ruslan Salakhutdinov.
\newblock Unsupervised learning of video representations using lstms.
\newblock In {\em ICML}, 2015.

\bibitem{RAFT}
Zachary Teed and Jia Deng.
\newblock {RAFT:} recurrent all-pairs field transforms for optical flow.
\newblock In Andrea Vedaldi, Horst Bischof, Thomas Brox, and Jan{-}Michael
  Frahm, editors, {\em ECCV}, 2020.

\bibitem{highfideity}
Ruben Villegas, Arkanath Pathak, Harini Kannan, Dumitru Erhan, Quoc~V. Le, and
  Honglak Lee.
\newblock High fidelity video prediction with large stochastic recurrent neural
  networks.
\newblock In {\em NeurIPS}, 2018.

\bibitem{villegas17mcnet}
Ruben Villegas, Jimei Yang, Seunghoon Hong, Xunyu Lin, and Honglak Lee.
\newblock Decomposing motion and content for natural video sequence prediction.
\newblock In {\em ICLR}, 2017.

\bibitem{Wang2018}
Ting-Chun Wang, Ming-Yu Liu, Jun-Yan Zhu, Guilin Liu, Andrew Tao, Jan Kautz,
  and Bryan Catanzaro.
\newblock Video-to-video synthesis.
\newblock In {\em NeurIPS}, 2018.

\bibitem{msssim}
Zhou Wang, Eero~P. Simoncelli, and Alan~C. Bovik.
\newblock Multiscale structural similarity for image quality assessment.
\newblock In {\em The Thrity-Seventh Asilomar Conference on Signals, Systems \&
  Computers, 2003}, volume~2, pages 1398--1402. IEEE, 2003.

\bibitem{FVS}
Yue Wu, Rongrong Gao, Jaesik Park, and Qifeng Chen.
\newblock Future video synthesis with object motion prediction.
\newblock In {\em CVPR}, 2020.

\bibitem{qvi_nips19}
Xiangyu Xu, Li Siyao, Wenxiu Sun, Qian Yin, and Ming-Hsuan Yang.
\newblock Quadratic video interpolation.
\newblock In {\em NeurIPS}, 2019.

\bibitem{vimeo}
Tianfan Xue, Baian Chen, Jiajun Wu, Donglai Wei, and William~T Freeman.
\newblock Video enhancement with task-oriented flow.
\newblock {\em International Journal of Computer Vision (IJCV)},
  127(8):1106--1125, 2019.

\bibitem{lpips}
Richard Zhang, Phillip Isola, Alexei~A Efros, Eli Shechtman, and Oliver Wang.
\newblock The unreasonable effectiveness of deep features as a perceptual
  metric.
\newblock In {\em CVPR}, 2018.

\bibitem{penn_action}
Weiyu Zhang, Menglong Zhu, and Konstantinos Derpanis.
\newblock From actemes to action: A strongly-supervised representation for
  detailed action understanding.
\newblock In {\em ICCV}, 2013.

\end{thebibliography}
}

\end{document}